\documentclass[journal]{IEEEtran}
\usepackage{amsmath}
\usepackage{amsfonts} 
\usepackage{amssymb}

\usepackage{multirow}
\usepackage{graphbox}
\usepackage{pifont}
\usepackage{tikz}
\usetikzlibrary{chains, positioning, fit, shapes.geometric, arrows, calc, decorations.pathreplacing}

\usepackage{booktabs}
\usepackage{algorithm}
\usepackage{algpseudocode}
\usepackage{makecell}
\usepackage{hyperref}

\usepackage{cite}
\usepackage{color}

\newcommand{\etal}[1]{\emph{et al}. }
\newcommand{\xmark}{\ding{55}}%

\newcommand{\best}[1]{{\color{blue}{#1}}}
\newcommand{\secondbest}[1]{{\color{red}{#1}}}
\ifCLASSINFOpdf
\else
\fi
\usepackage[caption=false,font=footnotesize]{subfig}
\usepackage{url}
\begin{document}
\title{Face anonymization preserving facial expressions and photometric realism}

\author{Luigi Celona, Simone Bianco, Raimondo Schettini% <-this % stops a space
\thanks{Luigi Celona, Simone Bianco and Raimondo Schettini are with the Department of Informatics, Systems and Communication, University of Milano-Bicocca, 20126 Milan, Italy (e-mail: \{first\_name\}.\{last\_name\}@unimib.it).}% <-this % stops a space
\thanks{Manuscript received April 19, 2005; revised August 26, 2015.}}


\markboth{Journal of \LaTeX\ Class Files,~Vol.~14, No.~8, August~2015}%
{Shell \MakeLowercase{\textit{et al.}}: Bare Demo of IEEEtran.cls for IEEE Journals}



\maketitle

\begin{abstract}
The widespread sharing of face images on social media platforms and in large-scale datasets raises pressing privacy concerns, as biometric identifiers can be exploited without consent. Face anonymization seeks to generate realistic facial images that irreversibly conceal the subject’s identity while preserving their usefulness for downstream tasks. However, most existing generative approaches focus on identity removal and image realism, often neglecting facial expressions as well as photometric consistency—specifically attributes such as illumination and skin tone—that are critical for applications like relighting, color constancy, and medical or affective analysis. In this work, we propose a feature-preserving anonymization framework that extends DeepPrivacy by incorporating dense facial landmarks to better retain expressions, and by introducing lightweight post-processing modules that ensure consistency in lighting direction and skin color. We further establish evaluation metrics specifically designed to quantify expression fidelity, lighting consistency, and color preservation, complementing standard measures of image realism, pose accuracy, and re-identification resistance. Experiments on the CelebA-HQ dataset demonstrate that our method produces anonymized faces with improved realism and significantly higher fidelity in expression, illumination, and skin tone compared to state-of-the-art baselines. These results underscore the importance of feature-aware anonymization as a step toward more useful, fair, and trustworthy privacy-preserving facial data.
\end{abstract}


% \fi
\IEEEpeerreviewmaketitle

%\IEEEPARstart{T}{his} demo file is intended to serve as a ``starter file'' for IEEE journal papers produced under \LaTeX\ using IEEEtran.cls version 1.8b and later.
% (should never be an issue) I wish you the best of success.

%========================================================
\section{Introduction}

\IEEEPARstart{A}{nonymization} refers to the process of removing or altering personally identifiable information within data to safeguard individuals’ privacy. In computer vision, large collections of visual data are indispensable for the development of robust recognition and analysis systems. However, these datasets often include privacy-sensitive contents such as faces, license plates, or other biometric and contextual cues. The publication and reuse of such data raise ethical and legal concerns, amplified by recent regulatory frameworks. In particular, the General Data Protection Regulation (GDPR) in Europe~\cite{regulation2016regulation}, the California Consumer Privacy Act (CCPA) in the United States~\cite{illman2019california}, and the Act on the Protection of Personal Information (APPI) in Japan~\cite{personal2017act} require that personal data be either anonymized or used only with explicit user content. Since obtaining consent at scale is rarely feasible, reliable anonymization has become a prerequisite for the ethical and lawful use of visual data.

\begin{figure*}
    \centering
    \setlength{\tabcolsep}{2pt}
    \begin{tabular}{cccccc}
        \includegraphics[width=0.15\linewidth]{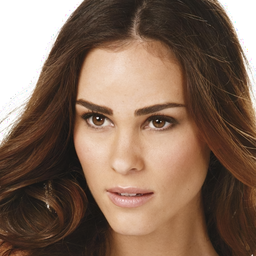} & 
        \includegraphics[width=0.15\linewidth]{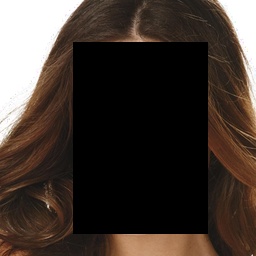} & 
        \includegraphics[width=0.15\linewidth]{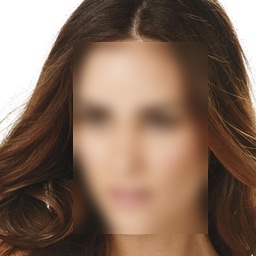} & 
        \includegraphics[width=0.15\linewidth]{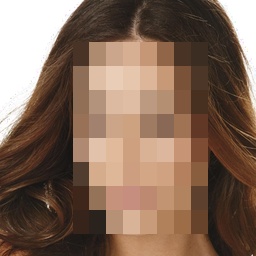} & 
        \includegraphics[width=0.15\linewidth]{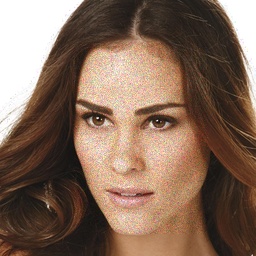} & 
        \includegraphics[width=0.15\linewidth]{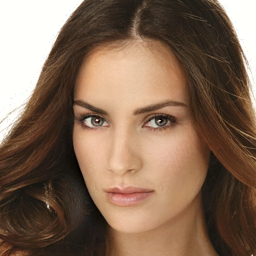} \\
        Original face & Masking & Blurring & Pixelization & Noise & Generation
    \end{tabular}
    \caption{Comparison of canonical image processing methods for face anonymization. From left to right are masking, blurring, pixelation,  noise, and generation \cite{hukkelaas2019deepprivacy}.}
    \label{fig:anony_samples}
\end{figure*}

Face anonymization methods can be broadly divided into two main categories: target-specific and target-generic approaches~\cite{sun2018hybrid}. Target-specific methods rely on adversarial perturbations to fool a particular recognition model~\cite{liang2021parallel,liang2022large,shan2020fawkes}. Although they can achieve near-perfect identity obfuscation for the attached model, they often fail to generalize to other recognition systems and may remain vulnerable to human identification. Target-generic methods, in contrast, modify the visual appearance of the face itself so that both humans and algorithms cannot infer the original identity. Early approaches in this direction—such as masking~\cite{yang2022study}, blurring~\cite{jiang2023dartblur}, or pixelation ~\cite{zhou2020personal}—are simple and effective but heavily compromise image utility. As shown in Figure \ref{fig:anony_samples}, these occlusion-based techniques disrupt structural and photometric information, making anonymized faces unsuitable for downstream tasks such as expression recognition, pose estimation, or relighting.

To overcome these limitations, learning-based approaches have emerged as a dominant paradigm. Generative Adversarial Network (GAN)-based methods synthesize artificial faces that preserve selected non-identity attributes, such as pose or approximate facial geometry, while removing identity-related features~\cite{hukkelaas2019deepprivacy,yang2024g,kung2024face}. For instance, DeepPrivacy~\cite{hukkelaas2019deepprivacy} introduced a conditional GAN framework to replace the original head region with an anonymized but realistic version, while CIAGAN~\cite{maximov2020ciagan} and IdentityDP~\cite{wen2022identitydp} extended this paradigm with more controllable identity vectors and privacy guarantees. More recent works have explored StyleGAN inversion~\cite{li2023riddle}, latent code optimization~\cite{barattin2023attribute}, and hybrid approaches combining adversarial perturbations with generative synthesis~\cite{hukkelaas2022deepprivacy2,cao2021personalized}. Diffusion-based anonymization has also emerged as a promising alternative, with methods such as Diff-Privacy~\cite{he2024diff} and LDFA~\cite{klemp2023ldfa} showing strong results in identity removal.

Despite these advances, the majority of existing methods focus primarily on preventing re-identification and maximizing image realism, often at the expense of photometric consistency—the accurate preservation of lighting and color characteristics across the original and anonymized face. This consistency is vital for downstream tasks where visual cues like illumination and skin tone must remain stable to avoid performance degradation or bias. Such tasks include:

\begin{itemize}
    \item Facial expression analysis, where anonymized data should retain the underlying emotion to remain useful for affective computing.
    \item Illuminant estimation and relighting, which require accurate preservation of lighting direction and intensity for tasks such as image enhancement or virtual reality rendering.
    \item Color constancy and fairness-aware recognition, where consistent skin tone reproduction ensures both technical reliability and social inclusivity.
\end{itemize}

Existing anonymization methods often distort shadows, alter shading, or shift skin tones, leading to anonymized datasets that are less representative and less useful for these tasks. In particular, landmark-conditioned GAN approaches frequently fail to preserve non-identity cues such as pose and gaze, which reduces the utility of the anonymized data for downstream applications.

In this work, we propose a feature-preserving face anonymization framework designed to address this gap. Building upon the DeepPrivacy architecture~\cite{hukkelaas2019deepprivacy}, we introduce architectural and photometric refinements that allow the preservation of three underrepresented but crucial features: facial expression, lighting direction, and skin color. Specifically, our contributions are as follows:

\begin{itemize}
    \item We enhance DeepPrivacy’s conditioning mechanism by providing a denser set of facial landmarks provided to the generator. This modification improves the retention of subtle expression details, resulting in anonymized faces that more faithfully reflect the affective state of the original subject.
    \item We design lightweight post-processing methods to reintroduce low-level image information into the anonymized output. A Laplacian pyramid blending technique transfers illumination cues from the input to the anonymized face, while a color transfer method in the YCbCr color space aligns skin tone statistics, ensuring that skin color is preserved without reintroducing identity.
    \item We consolidate existing metrics for realism, detection, pose preservation, and re-identification, and introduce new measures for expression retention, lighting consistency, and skin color fidelity. This evaluation framework provides a more complete understanding of the trade-offs in anonymization.
\end{itemize}

We validate our approach on the CelebA-HQ dataset, comparing against eight state-of-the-art anonymization methods. Experimental results demonstrate that our method improves upon existing baselines in terms of both qualitative visual realism and quantitative metrics, particularly in preserving expression, lighting, and skin tone. These improvements make anonymized images more useful for downstream vision tasks while maintaining privacy guarantees.

The remainder of this paper is organized as follows. In Section \ref{sec:related-work}, the related works are summarized. Section \ref{sec:method} presents the proposed method for face anonymization and the adopted evaluation metrics. Section~\ref{sec:experiments} details the experimental setup, including the datasets used, the benchmark methods considered for comparison, and the obtained results. Finally, Section~\ref{sec:conclusions} summarizes the key findings and presents concluding remarks.

%========================================================
\section{Related work}
\label{sec:related-work}
\subsection{Face Privacy}

\subsubsection{Face Anonymization}
Anonymization or de-identification methods concentrate on preserving facial attributes, such as gender, age, and race, while eliminating identity information from input images. Early face anonymization techniques mainly involve masking, pixelization, blurring, and similar approaches. However, these occlusion methods significantly diminish the data's usefulness (see Figure \ref{fig:anony_samples}). Additionally, they have proven to be ineffective for deep learning-based face recognition \cite{gross2006model}. An alternative approach involves $k$-same algorithm-based methods, wherein the average face of $k$-closest faces replaces the given face, reducing face recognition accuracy to less than 1/$k$ \cite{newton2005preserving}. Various modifications \cite{du2014garp,jourabloo2015attribute} aim to enhance the data utility of the average face. Recent advances incorporate new techniques, such as de-identification through adversarial perturbations \cite{liang2022large,shan2020fawkes,liang2021parallel}. Shan \etal~\cite{shan2020fawkes} proposed a system altering image feature representations using imperceptible perturbations. Generative Adversarial Networks (GANs) \cite{goodfellow2020generative} have inspired a new category of face de-identification techniques, divided into conditional inpainting-based \cite{sun2018natural,hukkelaas2019deepprivacy,hukkelaas2022deepprivacy2} and face representations manipulating-based \cite{gafni2019live,maximov2020ciagan,wen2022identitydp}. For example, Sun \etal~\cite{sun2018natural} generate realistic head inpainting based on facial landmarks, while DeepPrivacy \cite{hukkelaas2019deepprivacy} utilized a GAN-based head inpainting technique for comprehensive privacy removal. Gafni \etal~\cite{gafni2019live} generate high-level representations from face images, minimizing identity associations while retaining perceptions (pose, illumination, and expression). CIAGAN \cite{maximov2020ciagan} employs a vector to control the fake identity, and IdentityDP \cite{wen2022identitydp} combines differential privacy mechanisms with deep neural networks for adjustable privacy control. Cao \etal~\cite{cao2021personalized} introduced user-specific passwords and adjustable parameters for invertible de-identification. FALCO \cite{barattin2023attribute} optimizes latent representation in a pre-trained GAN's space, synthesizing face images for effective de-identification and attribute preservation. RiDDLE \cite{li2023riddle} involves mapping the original face image into the latent space of a pre-trained StyleGAN2. Subsequently, the latent code and password undergo processing using a streamlined transformer, ultimately producing encrypted code. Similarly, HFORD \cite{chen2023hford} achieves high-fidelity and occlusion-robust deidentification by adding an attribute retention module. Leveraging the advancements in diffusion models, He \etal~\cite{he2024diff} introduced Diff-Privacy, merging anonymization and identity information hiding via a multi-scale image inversion module. This approach utilizes a stable diffusion model for robust privacy protection. AIDPro~\cite{wang2025beyond} leverages a hybrid architecture combining a face anonymization GAN with a signature embedding network to ensure both identity concealment and verification capability.

\subsubsection{Identity-Preserving Face Privacy}
These methods aim to alleviate privacy concerns in face recognition systems by preventing the use of facial information beyond identity recognition. Existing methods fall into two categories: image-level and representation-level. At the image level, Othman \etal~\cite{othman2015privacy} introduced the concept of soft biometric privacy, proposing a face-mixing approach that conceals gender attributes while preserving recognition utility. Adversarial perturbation-based methods \cite{mirjalili2018semi,mirjalili2017soft} are introduced for further enhancement. Chhabera \etal~\cite{chhabra2018anonymizing} applied adversarial perturbation for k-facial attributes anonymization, and Mirjalili \etal~extended semi-adversarial networks for multi-attribute privacy in facial images \cite{mirjalili2020privacynet}. Li \etal~\cite{li2021identity,li2021learning} proposed visual appearance obfuscation methods for face image protection. Although these methods prevent automatic mining of facial attributes, some results are model-dependent and suffer from artifacts. Minor visual changes may not guarantee privacy concerning human observers, leading to risks like malicious AI face swapping without consent. Our work introduces a high-resolution face obfuscation framework to address these issues. At the representation level, some researchers focused on privacy in face descriptors \cite{morales2020sensitivenets,terhorst2021soft,terhorst2019unsupervised}, aiming for facial representations without privacy-sensitive attributes. For instance, SensitiveNet~\cite{morales2020sensitivenets} eliminates sensitive information using a modified triplet loss. Terhörst \etal~\cite{terhorst2019unsupervised,terhorst2021soft} proposed methods to suppress privacy-sensitive information on the template level. However, these methods cannot protect images in face recognition systems and heavily depend on face descriptor extraction models. Our proposal introduces a privacy-preserving framework for face images rather than representations.

\subsection{Anonymization Metrics}
Although many face anonymization methods have been proposed in the literature, their evaluation is often just partial since many aspects have to be taken into account. This is why the first attempts towards a comprehensive evaluation methodology have recently started to appear \cite{hanisch2024false,cao2024face}.

Face anonymization methods are typically evaluated based on two primary criteria: the effectiveness of privacy protection and the visual quality and/or utility of the generated faces.

Among the privacy preserving metrics, the principal one is the reduction in recognition accuracy. The metric evaluates how well an anonymization method obscures an individual's identity by measuring the decrease in accuracy of facial recognition systems when applied to anonymized images. A similar metric exploits the computation of face identity embeddings, a step used by almost all face verification models. To determine whether two images are of the same identity, they comparing the distance between their identity embeddings; therefore an increase in the distance between the original and anonymized face embeddings indicates an effective anonymization \cite{cao2024face}. Todt \etal~\cite{todt2022fant} investigated the phenomenon of face anonymization reversibility, testing the potential to reconstruct or reverse the anonymized image back to its original form. Techniques that resist such reconstruction attempts are considered more robust in preserving privacy.

Visual quality and utility preserving metrics  focus on measuring the naturalness and quality of the generated images, or their utility for downstream computer vision tasks \cite{stenger2024evaluating}. This has been done measuring for example face detectability, i.e. how well anonymized faces are detected by face detectors \cite{klomp2021safe,cao2024face}. Sometimes this metric is complemented with a more fine-grained metric to measure the distance of face landmarks before and after anonymization, to evaluate to what extent their position has been preserved, or even if non-biometric facial attributes have been preserved  \cite{cai2024disguise}.
From a visual quality point of view, common metrics taken from the image quality assessment field are usually used. This type of features comprises signal-based metrics, as for example the Peak signal-to-noise ratio (PSNR) or the Structural Similarity index (SSIM) \cite{wang2004image}, as well as more semantic-aware metrics, such as, for example the Fr\'echet Inception Distance (FID) \cite{heusel2017gans}, that measures the distance between the distribution of original and anonymized face images, with a low FID indicating a high similarity. 

%========================================================

\section{Proposed Method}
\label{sec:method}

In this section, we first formalize the face anonymization problem. We then describe our modifications to the existing GAN-based anonymization framework, followed by the evaluation methodology.

\subsection{Problem Formalization}
Let $I_o \in \mathbb{R}^{H \times W \times 3}$ denote the original face image with identity $y \in \mathcal{Y}$, where $\mathcal{Y}$ is the set of possible identities. The goal of face anonymization is to generate a new image $I_a$ that conceals the identity $y$ while retaining useful visual information.

Formally, let $\mathcal{A}$ be the anonymization function. The anonymized image is defined as:
\[
I_a = \mathcal{A}(I_o),
\]
where $I_a \in \mathbb{R}^{H \times W \times 3}$ is the output image that no longer reveals the subject's identity but preserves key non-identity features. An effective anonymization method must satisfy the following criteria:

\begin{enumerate}
    \item Privacy preservation. The probability of re-identifying $y$ from $I_a$ is minimized. Formally, for a face recognition model $f:\mathbb{R}^{H \times W \times 3}\to\mathcal{Y}$,
    \[
    \Pr[f(I_a) = y] \approx 0.
    \]
    \item Utility preservation. The anonymized image should retain non-identity features such as pose, facial expression, illumination, and skin tone. Let $\phi_k(\cdot)$ denote attribute extractors for task $k$ (e.g., pose, expression). We require
    \[
    d\big(\phi_k(I_o), \phi_k(I_a)\big) \leq \epsilon_k,
    \]
    where $d(\cdot, \cdot)$ is a task-specific distance and $\epsilon_k$ a tolerance threshold. In our framework, utility preservation specifically includes the retention of photometric attributes such as illumination direction and skin tone—collectively referred to as photometric consistency. These low-level features are essential for maintaining image utility in applications where lighting cues and color constancy matter.
    \item Realism. The output $I_a$ should resemble a natural image, consistent with the distribution of real facial data. This is typically enforced by generative models and quantified by metrics such as the Fr\'echet Inception Distance (FID)~\cite{heusel2017gans}.
\end{enumerate}

\subsection{Anonymization Method}
Our approach builds upon the DeepPrivacy framework~\cite{hukkelaas2019deepprivacy}, which employs a conditional GAN-based inpainting strategy to replace the head region with a synthetic but realistic alternative. We extend this framework with two main modifications.

\subsubsection{Landmarks}
\begin{figure}
    \centering
    \begin{tabular}{cc}
        \includegraphics[width=0.45\columnwidth]{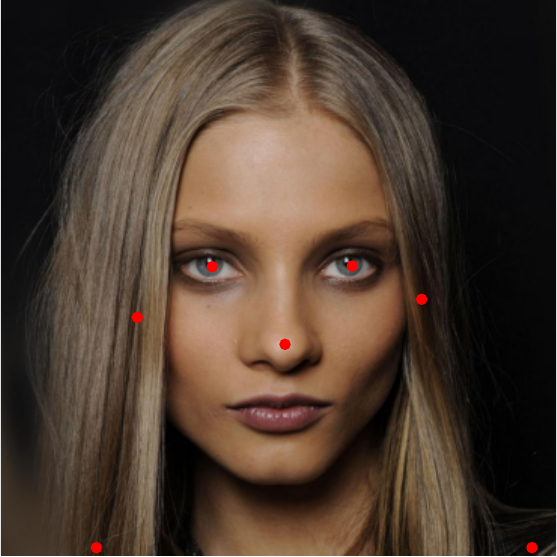} & \includegraphics[width=0.45\columnwidth]{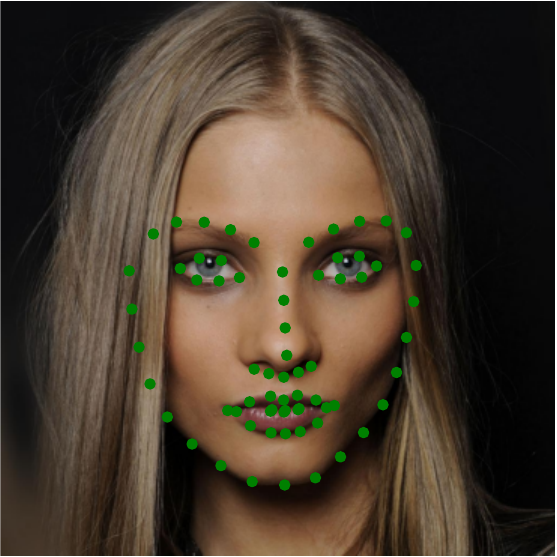} \\
         (a) & (b)
    \end{tabular}
    \caption{Comparison of pose-conditioning landmarks for anonymization. (a) Sparse set of seven landmarks extracted with Mask R-CNN~\cite{he2017mask}, covering only ears, eyes, nose, and shoulders as in the original DeepPrivacy framework. (b) Dense set of 68 facial landmarks detected with the DLib implementation of~\cite{kazemi2014one}, providing finer geometric detail and enabling more faithful expression preservation.}
    \label{fig:landmarks-deepprivacy-vs-our}
\end{figure}

DeepPrivacy conditions its anonymized face synthesis on pose information extracted during decoding. In the original framework, sparse pose estimation is performed using Mask R-CNN~\cite{he2017mask}, providing only seven landmarks corresponding to the ears, eyes, nose, and shoulders. In our approach, we replace this sparse representation with a denser landmark configuration consisting of 68 facial keypoints (see Fig.~\ref{fig:landmarks-deepprivacy-vs-our}). For this purpose, we employ the implementation of Kazemi and Sullivan~\cite{kazemi2014one} available in the DLib library~\cite{king2009dlib}. By enriching the conditioning space with a more detailed geometric prior, our model is able to capture subtle facial deformations and expression cues, thereby improving the perceptual fidelity of the anonymized output. Given that this change impacts the information encoded and processed by the model, we need to retrain DeepPrivacy by changing the pose information given in input.

\subsubsection{Post-Processing for Relighting and Color Transfer}

After the anonymized face $I_a = \mathcal{A}(I_o)$ is generated by the GAN, we apply a two-step post-processing procedure to restore the photometric properties of the original image $I_o$: (i) relighting based on shading transfer and (ii) color transfer to preserve skin tone consistency.

\paragraph{Intrinsic decomposition for relighting}
We assume that a face image $I$ can be factorized according to image intrinsic decomposition \cite{barrow1978recovering} into an albedo $A$ and a shading map $S$ as:
\[I(x,y) \approx A(x,y) \cdot S(x,y),\]
where $A(x,y) \in [0,1]^3$ encodes reflectance and $S(x,y) \in [0,1]$ encodes illumination intensity.

An approximate decomposition is obtained using Retinex-style bilateral filtering~\cite{jeon2014intrinsic}:
\begin{align}
S &= \mathcal{B}\!\big(\mathrm{gray}(I)\big), \\
A &= \frac{I}{S},
\end{align}
where $\mathcal{B}(\cdot)$ denotes a bilateral filter applied to the grayscale image.

We adopt intrinsic image decomposition as the core mechanism for relighting because it explicitly separates illumination and reflectance components, allowing the transfer of lighting information from the original to the anonymized face without reintroducing identity cues encoded in the albedo. In contrast to learned relighting networks such as SfSNet~\cite{sengupta2018sfsnet}, intrinsic decomposition is model-agnostic, computationally lightweight, and does not require additional training data. This makes it particularly suitable for anonymization frameworks, where photometric correction must be efficient and interpretable, and where maintaining privacy precludes reliance on large-scale face-specific relighting datasets. Moreover, this physically motivated formulation ensures consistent illumination transfer under varying imaging conditions, aligning with our goal of preserving photometric consistency across anonymized faces in a resource-efficient manner.

Given the original image $I_o$ and the anonymized output $I_a$, relighting is performed by combining the albedo of the anonymized face with the shading of the original:
\[I_{\text{relight}} = A_{I_a} \cdot S_{I_o}.\]

\begin{figure*}
    \centering
    \includegraphics[width=.8\textwidth]{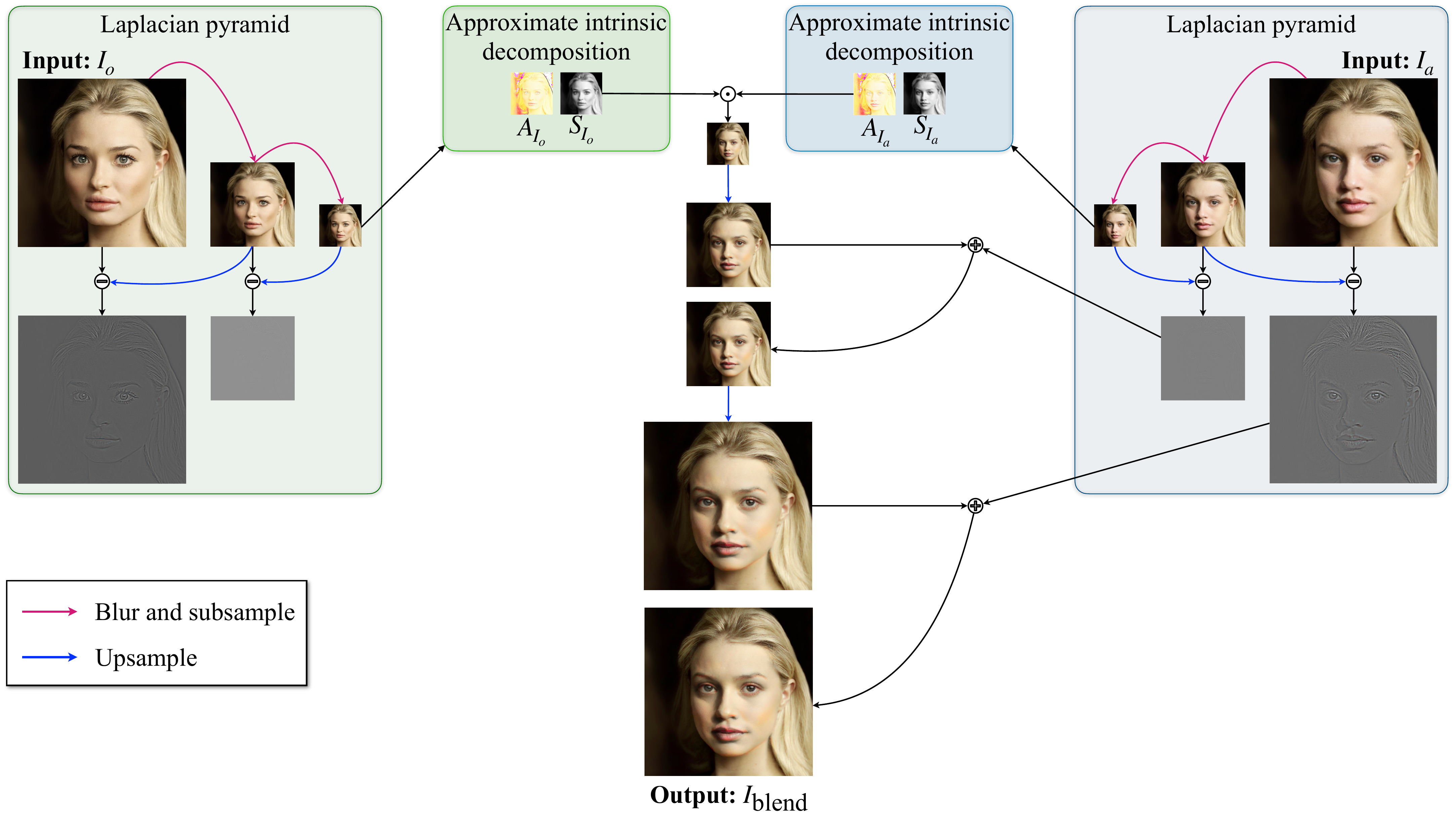}
    \caption{Relighting pipeline. The original image $I_o$ and anonymized image $I_a$ are decomposed into albedo ($A$) and shading ($S$) components. The shading of $I_o$ is combined with the albedo of $I_a$, and a Laplacian pyramid is used to refine consistency across spatial frequencies, producing the blended output $I_{\text{blend}}$.}
    \label{fig:relighting-pipeline}
\end{figure*}

As illustrated in Figure~\ref{fig:relighting-pipeline}, to improve consistency across spatial frequencies, we refine relighting with a Laplacian pyramid. For each level $k$, the Laplacian residual is
\[L^k = G^k - \text{upsample}(G^{k+1}),\]
where $G^k$ is the Gaussian pyramid. We substitute the low-frequency $I_a$ with the $I_{\text{relight}}$, then reconstruct the blended image:
\[I_{\text{blend}} = \sum_{k=1}^K \text{upsample}(L^k).\]

\paragraph{Color transfer}

To ensure that the anonymized output preserves the skin tone of the original subject, we perform a channel-wise color transfer. Specifically, we align the chromatic statistics of the blended image $I_{\text{blend}}$ with those of the original image $I_o$. The final corrected image is obtained as
\[
I_f = \mathcal{C}\big(I_{\text{blend}}, I_o\big),
\]
where $\mathcal{C}(\cdot)$ denotes the color transfer operator. 

Formally, let $\mu_t^c$ and $\sigma_t^c$ denote the mean and standard deviation of channel $c$ in $I_{\text{blend}}$, and $\mu_r^c$, $\sigma_r^c$ the corresponding statistics in $I_o$. For each pixel $(x,y)$ and channel $c$, we define:
\[
I_f^c(x,y) = \frac{\sigma_r^c}{\sigma_t^c}\left(I_{\text{blend}}^c(x,y) - \mu_t^c\right) + \mu_r^c.
\]
Here, the channels $c$ correspond to the chrominance components ($Cb$, $Cr$) of the YCbCr color space, while the luminance channel ($Y$) is left unchanged to avoid altering structural information. After transformation, pixel values are clipped to the valid range $[0, 255]$ and converted back to the RGB space.

%This procedure yields a final anonymized image $I_f$ that inherits the shading properties of $I_o$ while maintaining its original skin tone distribution, thereby enhancing both visual fidelity and fairness in downstream applications.

\subsection{Anonymization Performance Evaluation}
To evaluate anonymization quality, we compare multiple generative models under a unified set of criteria. Our evaluation framework considers three complementary dimensions:
\begin{itemize}
    \item Image quality -- measuring perceptual realism and fidelity of the generated images.
    \item Privacy -- quantifying the ability of the anonymization method to prevent identity leakage.
    \item Utility -- assessing whether the anonymized images remain usable for downstream computer vision tasks, such as pose estimation, expression analysis, or relighting.
\end{itemize}

To provide a comprehensive analysis, we report both commonly used metrics from the literature and our proposed metrics tailored to specific utility aspects not fully captured by standard benchmarks.

\subsubsection{Common Evaluation Metrics}

The performance of face anonymization models is typically assessed through four widely adopted criteria, namely visual quality, face detection rate, pose preservation, and face re-identification risk \cite{hukkelaas2019deepprivacy,maximov2020ciagan,dall2022graph}. \textbf{Visual quality} is quantified using the Fr\'echet Inception Distance (FID) \cite{heusel2017gans}, which measures the distance between distributions of real and anonymized images in a deep feature space. These features are extracted from specific layers of an Inception network pre-trained on ImageNet, yielding FID${\textrm{ImageNet}}$. Since FID${\textrm{ImageNet}}$ may be biased by the fact that many generative models are trained on ImageNet-pretrained backbones, we also compute a CLIP-based variant, FID$_{\textrm{CLIP}}$ \cite{kynkaanniemi2022role}, which leverages the more semantically aligned CLIP embedding space. Lower FID scores indicate higher image realism and closer alignment with the distribution of real faces. The \textbf{detection rate} is defined as the percentage of anonymized images in which a valid face is detected relative to the total number of images. This metric serves as a proxy for whether the generated faces remain recognizable by generic face detectors, a necessary condition for downstream use. We employ the MTCNN face detector \cite{zhang2016joint} for this task, and an ideal anonymization method should approach a detection rate of 100\%. \textbf{Pose preservation} ensures that anonymized images retain the geometric arrangement of facial landmarks. It is measured by the L$_1$ distance between the 68 landmarks detected on the original and anonymized faces, normalized by the inter-ocular distance \cite{kazemi2014one}. We use the DLib implementation \cite{king2009dlib} for landmark detection, which captures facial contours, eyes, mouth corners, and eyebrows. Finally, we assess \textbf{face re-identification risk} by computing the cosine similarity between identity embeddings of original and anonymized faces. Since CelebA-HQ does not provide paired identity labels, we apply anonymization to each test sample and extract identity embeddings using two variants of FaceNet \cite{schroff2015facenet}, one trained on VGGFace2 \cite{cao2018vggface2} and another on CASIA-WebFace \cite{yi2014learning}. Lower similarity scores indicate better identity obfuscation.

\subsection{Proposed utility metrics}

To complement the common metrics and better capture the utility dimension, we introduce additional metrics targeting expression, lighting, and color consistency. The \textbf{facial expression coherence} between the original and anonymized faces is measured using two complementary approaches. First, we compute the normalized L$_1$ distance between 51 expression-related facial landmarks (excluding contour points), focusing on regions such as eyes, eyebrows, nose, and mouth. Lower distances indicate better expression preservation. Second, we assess affective consistency by evaluating the accuracy of a pretrained emotion classifier in predicting the same emotion label for both the original and anonymized faces \cite{zhang2024leave}. Following the relighting evaluation framework proposed in \cite{zhou2019deep}, we quantify \textbf{lighting direction error} using the scale-invariant Mean Squared Error (Si-MSE) and scale-invariant L$_2$ norm (Si-L$_2$) \cite{barron2014shape} between the original RGB face image $\mathbf{I}_o$ and its anonymized counterpart $\mathbf{I}_a$:

\begin{equation}
\textrm{Si-MSE} = \frac{1}{N_I} \min_{\alpha}\lVert\mathbf{I}_o - \alpha \mathbf{I}_a\rVert^2_2,
\end{equation}

where $N_I$ is the number of facial pixels and $\alpha$ is a scalar that accounts for global intensity differences. The use of $\alpha$ ensures invariance to the overall brightness scaling while remaining sensitive to illuminant color. To further refine the lighting evaluation, we compute the scale-invariant L$_2$ distance between the estimated shadings $\mathbf{S}_o$ and $\mathbf{S}_a$:

\begin{equation}
\textrm{Si-L$2$} = \min{\alpha}\sqrt{\lVert\mathbf{S}_o - \alpha \mathbf{S}_a\rVert^2_2}.
\end{equation}

\begin{figure}
    \centering
    \setlength{\tabcolsep}{1pt}
    \begin{tabular}{cccc}
    \includegraphics[width=.24\columnwidth]{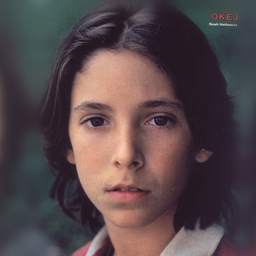} & \includegraphics[width=.24\columnwidth]{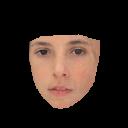} & \includegraphics[width=.24\columnwidth]{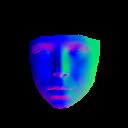} & \includegraphics[width=.24\columnwidth]{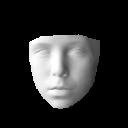} \\
        Face image & Albedo & Normal & Shading
    \end{tabular}
    \caption{Albedo, normal and shading images estimated for a face image by the SfSNet model \cite{sengupta2018sfsnet}.}
    \label{fig:sfsnet_output}
\end{figure}

Shading images are extracted using SfSNet \cite{sengupta2018sfsnet}, which also provides albedo and normal maps (see Fig.~\ref{fig:sfsnet_output}).
Finally, to measure \textbf{color similarity}, we compare the albedo images of the original and anonymized faces in the $L^*a^*b^*$ color space. We compute per-pixel lightness difference $\Delta L$, chroma difference $\Delta C$, and perceptual color difference $\Delta E_{2000}$ using the implementation from the \texttt{colour-science} library \cite{mansencal_thomas_2022_6288658} as follows:
\begin{equation}
\begin{aligned}
    &\Delta L = \mid L^*_o - L^*_a \mid, \\
    &\Delta C = \mid \sqrt{(a^*_o)^2 + (b^*_o)^2} -  \sqrt{(a^*_a)^2 + (b^*_a)^2} \mid,
\end{aligned}
\end{equation}

where $L^*_o, a^*_o, b^*_o$ are the original image lightness and chroma values and $L^*_a, a^*_a, b^*_a$ are the generated image lightness and chroma values. The mean values of these differences serve as image-level descriptors, indicating how closely the anonymized face reproduces the original’s intrinsic color appearance.

\subsection{Discussion}

\begin{table*}
\centering
\caption{Summary of Evaluation Metrics Used for Anonymization Performance Assessment. Metrics are grouped by evaluation dimension, with their purpose and computation method explicitly stated.}
\label{tab:evaluation_metrics}
\begin{tabular}{p{2.3cm} p{2.5cm} p{12cm}}
\toprule
Dimension & Metric & Description / Computation \\
\midrule
\multirow{3}{*}{Image Quality}
& FID${\textrm{ImageNet}}$ & Fr\'echet Inception Distance computed in ImageNet-pretrained Inception feature space \cite{heusel2017gans}. Lower is better. \\
& FID${\textrm{CLIP}}$ & FID computed in CLIP embedding space for semantically aligned evaluation \cite{kynkaanniemi2022role}. \\
& Detection Rate & Percentage of anonymized images where MTCNN detects a valid face \cite{zhang2016joint}. Higher is better. \\
\midrule
\multirow{2}{*}{Privacy}
& Cosine Similarity of Identity Embeddings & Cosine similarity between FaceNet embeddings of original and anonymized images, using two FaceNet versions (VGGFace2 and CASIA-WebFace) \cite{schroff2015facenet,cao2018vggface2,yi2014learning}. Lower indicates better anonymization. \\
& Face Re-identification Rate & Percentage of original-anonymized pairs matched by a re-identification network above a similarity threshold. \\
\midrule
\multirow{2}{*}{Geometry}
& Pose Preservation & Normalized L$_1$ distance between 68 facial landmarks detected by DLib \cite{kazemi2014one,king2009dlib}. Lower is better. \\
& Expression Coherence & (1) Normalized L$_1$ distance between 51 expression-related landmarks; (2) Emotion prediction consistency using a pretrained emotion classifier \cite{zhang2024leave}. \\
\midrule
\multirow{2}{*}{Lighting}
& Si-MSE & Scale-invariant Mean Squared Error between original and anonymized facial regions \cite{barron2014shape,zhou2019deep}. \\
& Si-L$_2$ & Scale-invariant L$2$ distance between shading maps estimated by SfSNet \cite{sengupta2018sfsnet}. \\
\midrule
Color & $\Delta L$, $\Delta C$, $\Delta E{2000}$ & Pixelwise lightness, chroma, and perceptual color differences computed in $L^*a^*b^*$ color space using \texttt{colour-science} \cite{mansencal_thomas_2022_6288658}. Lower is better. \\
\bottomrule
\end{tabular}
\end{table*}

By combining metrics from multiple dimensions—image quality, privacy, and utility—we ensure that the evaluation does not overfit to a single objective (e.g., identity suppression) at the expense of others (e.g., expression fidelity). This holistic approach provides a more reliable picture of an anonymization method’s real-world applicability, where generated faces must be visually convincing, privacy-preserving, and still usable for downstream tasks such as expression analysis, pose estimation, or lighting-aware applications.

\section{Experiments}
\label{sec:experiments}
This section describes the datasets, evaluation metrics, state-of-the-art baselines, and implementation details used to benchmark our approach. The goal is to provide a reproducible and rigorous assessment of the proposed anonymization method.

\subsection{Datasets}

We use the Flickr-Faces-HQ (FFHQ) dataset \cite{karras2019style} to train our model and CelebA-HQ \cite{karras2017progressive} for evaluation.

FFHQ consists of 70,000 high-resolution images at $1024 \times 1024$ pixels, exhibiting substantial diversity in age, ethnicity, and background, as well as accessories such as eyeglasses, sunglasses, hats, and facial hair. The images were sourced from Flickr, inheriting its demographic and contextual biases, and have been automatically aligned and cropped.

CelebA-HQ is a high-quality subset of the CelebA dataset \cite{liu2015faceattributes}, containing 30,000 celebrity face images at $1024 \times 1024$ resolution. Each image is annotated with 40 attribute labels describing facial properties (e.g., smiling, wearing glasses, hair color), which are particularly valuable for assessing identity obfuscation and attribute preservation.

%FFHQ \cite{karras2019style} is used to train our model, while CelebA-HQ \cite{karras2017progressive} is used for evaluation. FFHQ consists of 70,000 high-resolution images at 1024 × 1024 pixels, FFHQ showcases significant diversity in age, ethnicity, and background. It includes various accessories such as eyeglasses, sunglasses, hats, etc. The images were crawled from Flickr, thus they inherit the biases of the platform and undergo automatic alignment and cropping. CelebA-HQ comprises 30,000 face images of celebrities sourced from the CelebA dataset \cite{liu2015faceattributes}. These images have a resolution of 1024 × 1024 pixels and feature various demographic attributes such as age, gender, and race, and each image is annotated with 40 attribute labels related to the inner and outer regions of the face.

\subsection{Evaluation metrics}
%We evaluate our method by quantifying image quality, pose preservation, privacy preservation, facial expression preservation, skin color preservation, and lighting direction preservation. For convenience, all metrics used in this work are summarized in Table~\ref{tab:evaluation_metrics}, grouped by evaluation dimension and accompanied by their computation methods.
To rigorously assess anonymization performance, we evaluate our method across six key dimensions: image quality, pose preservation, privacy preservation, facial expression preservation, skin color preservation, and lighting direction preservation. The evaluation protocol follows a unified, reproducible methodology consistent with prior works \cite{hukkelaas2019deepprivacy,maximov2020ciagan,dall2022graph}, ensuring fair comparison across different models. For clarity, all metrics used in this work are summarized in Table~\ref{tab:evaluation_metrics}, grouped by evaluation dimension (image quality, privacy, and utility) and accompanied by their computation methods.

\subsection{State-of-the-art methods}
We benchmark our approach against eight state-of-the-art face anonymization methods representing different modeling strategies:

\begin{itemize}
    \item AIDPro \cite{wang2025beyond}: a recent method focusing on identity-disentangled face generation with strong privacy guarantees.
    \item AnonyGAN \cite{dall2022graph}: a GAN-based anonymization model incorporating pose conditioning.
    \item CIAGAN \cite{maximov2020ciagan}: a conditional identity-anonymizing GAN leveraging pose heatmaps for identity removal.
    \item DeepPrivacy \cite{hukkelaas2019deepprivacy} and DeepPrivacy2 \cite{hukkelaas2022deepprivacy2}: privacy-preserving generative models using bounding-box-based anonymization, with DeepPrivacy2 introducing style-based generators.
    \item FALCO \cite{barattin2023attribute}: an attribute-controllable anonymization framework that explicitly preserves non-identity attributes.
    \item G$^2$Face \cite{yang2024g}: a geometry-guided anonymization approach designed to maintain pose and expression fidelity.
    \item RiDDLE \cite{li2023riddle}: a disentanglement-based method separating identity and attribute representations for controllable anonymization.
\end{itemize}

We use the official implementations and pretrained weights provided by the authors where available, or faithfully re-implement the methods following their original papers when code is unavailable. All baselines are evaluated under the same resolution, preprocessing pipeline, and test split to ensure a fair and direct comparison.  

%AnonyGAN~\cite{dall2022graph}, CIAGAN~\cite{maximov2020ciagan}, DeepPrivacy~\cite{hukkelaas2019deepprivacy}, DeepPrivacy2~\cite{hukkelaas2022deepprivacy2}, FALCO~\cite{barattin2023attribute}, RiDDLE~\cite{li2023riddle}, G$^2$Face \cite{yang2024g}, and AIDPro \cite{wang2025beyond}.

\subsection{Results}
\label{sec:results}

\begin{figure}
    \setlength{\tabcolsep}{1pt}
    \begin{tabular}{cccc}
         Original & \includegraphics[align=c,width=.12\textwidth]{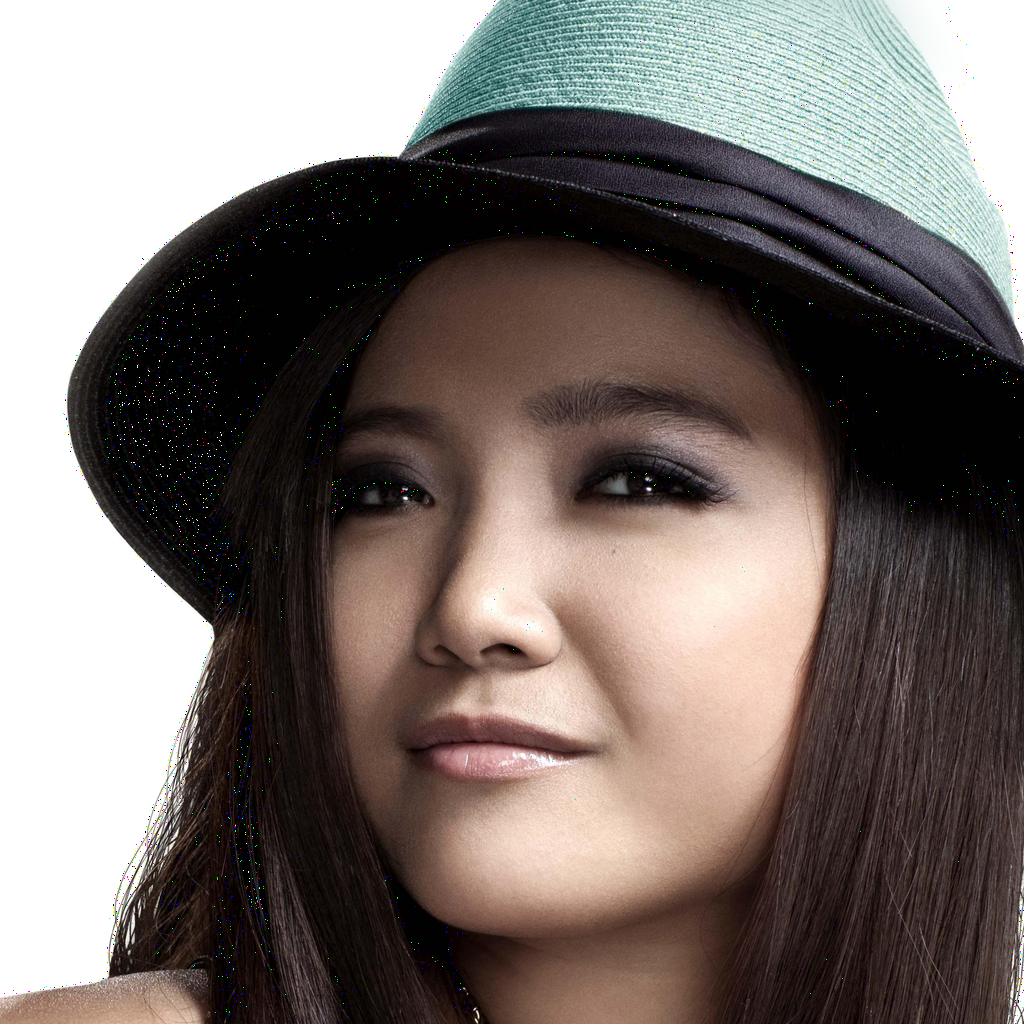} & \includegraphics[align=c,width=.12\textwidth]{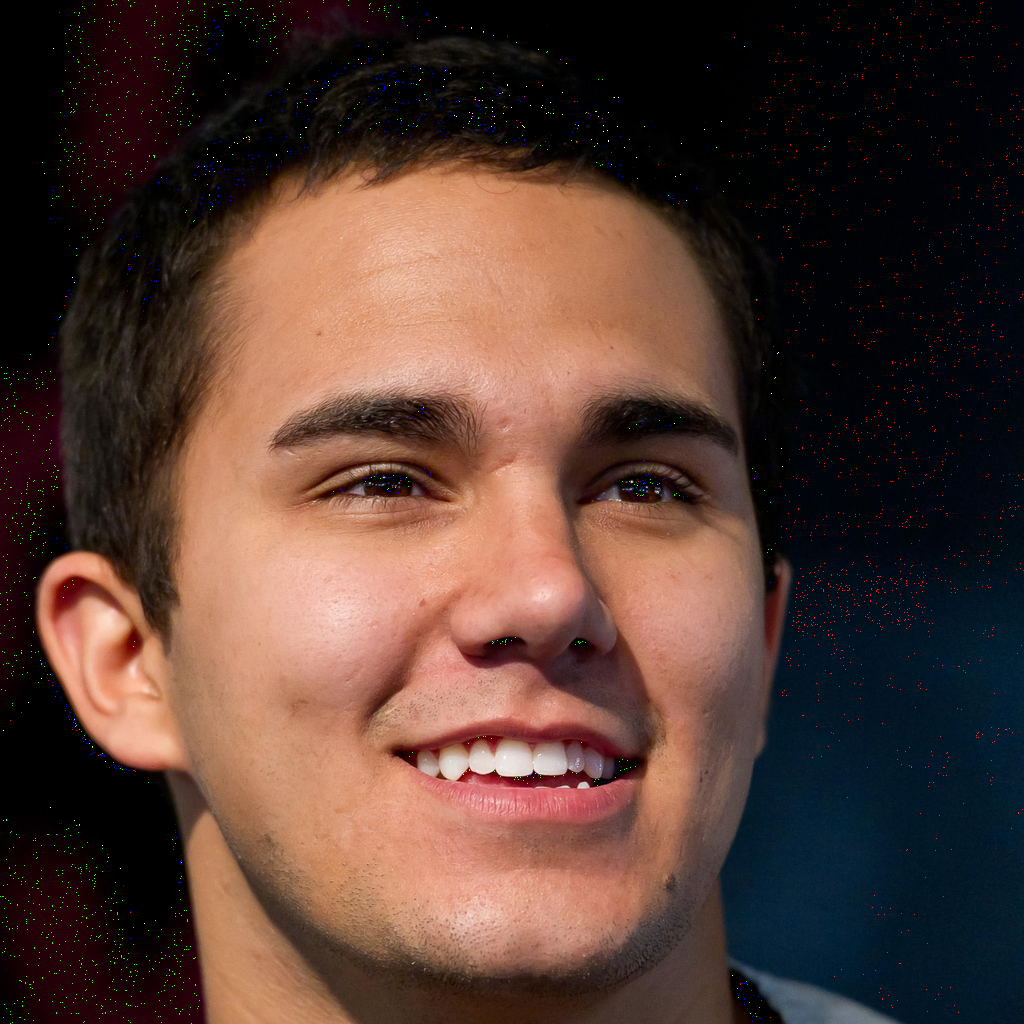} & \includegraphics[align=c,width=.12\textwidth]{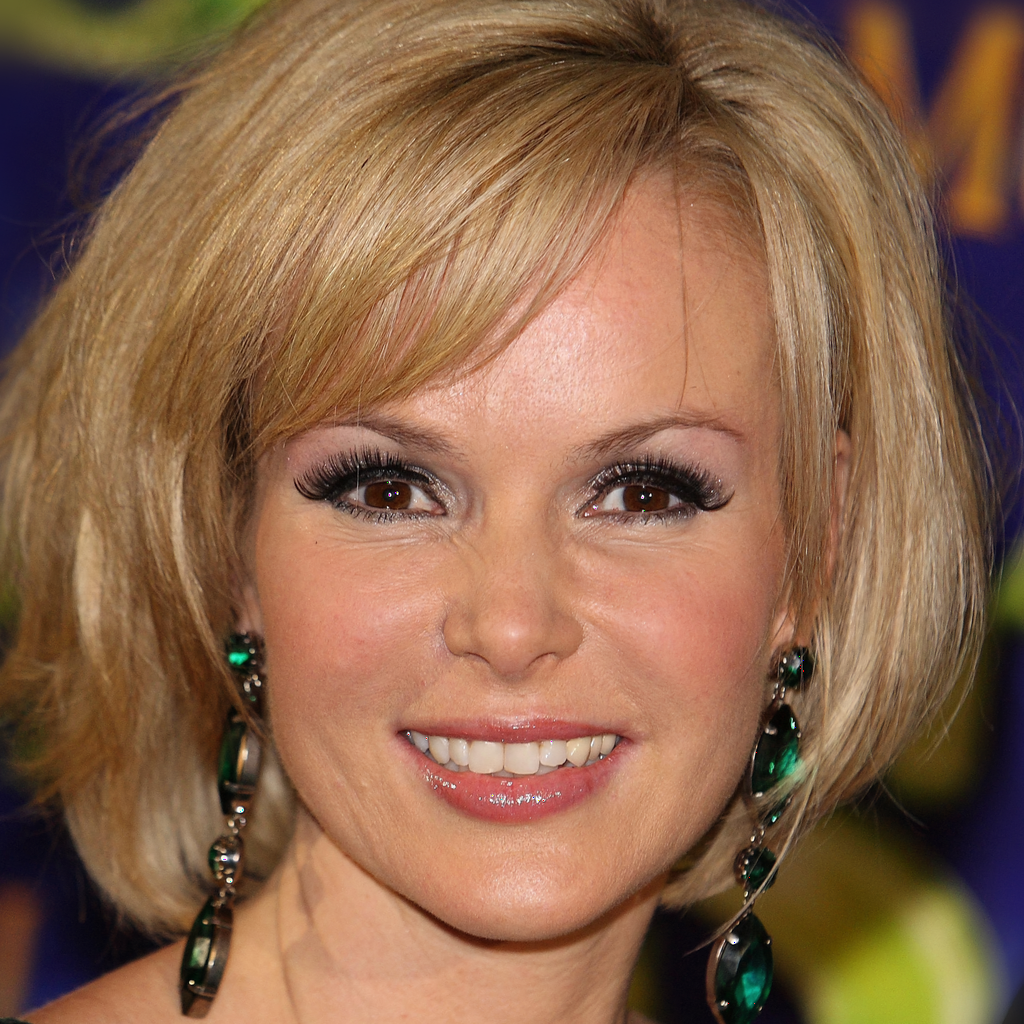} \vspace{.2em}\\
         AIDPro & \includegraphics[align=c,width=.12\textwidth]{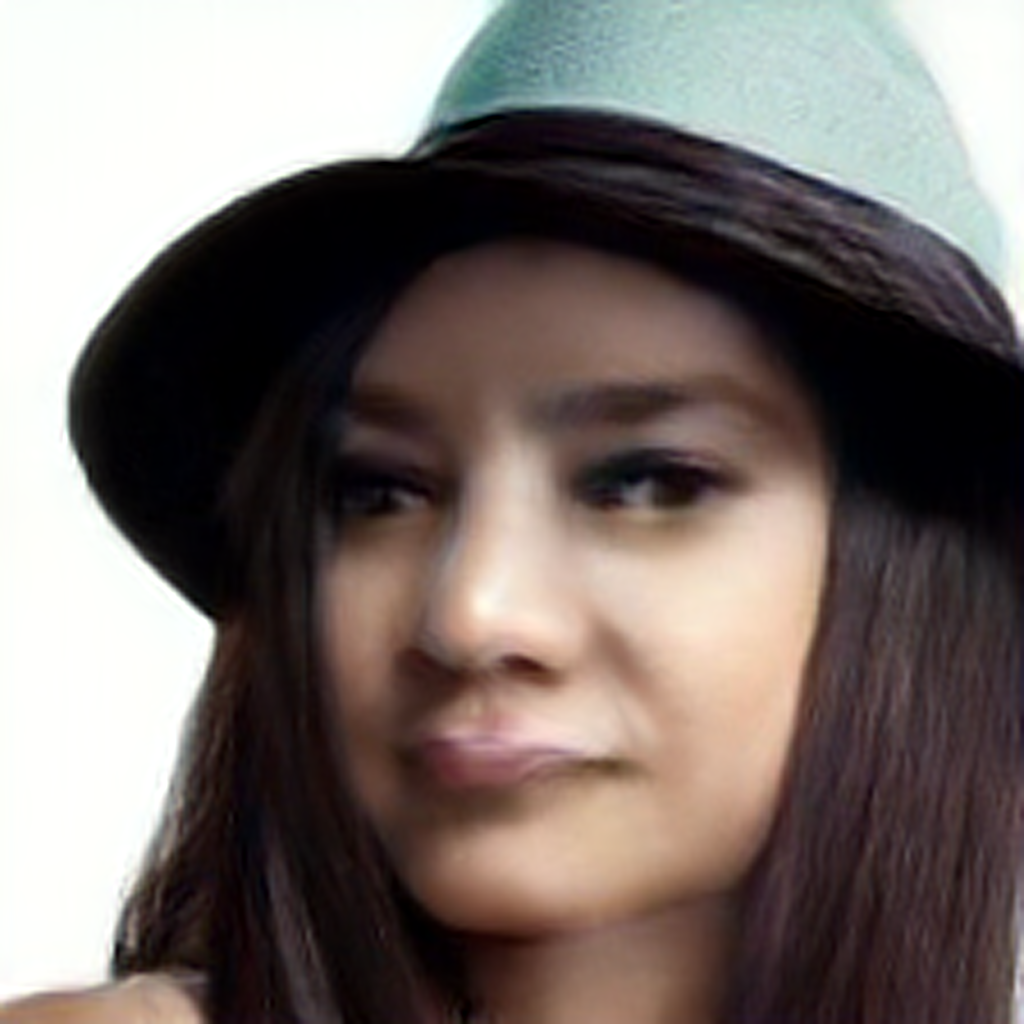} & \includegraphics[align=c,width=.12\textwidth]{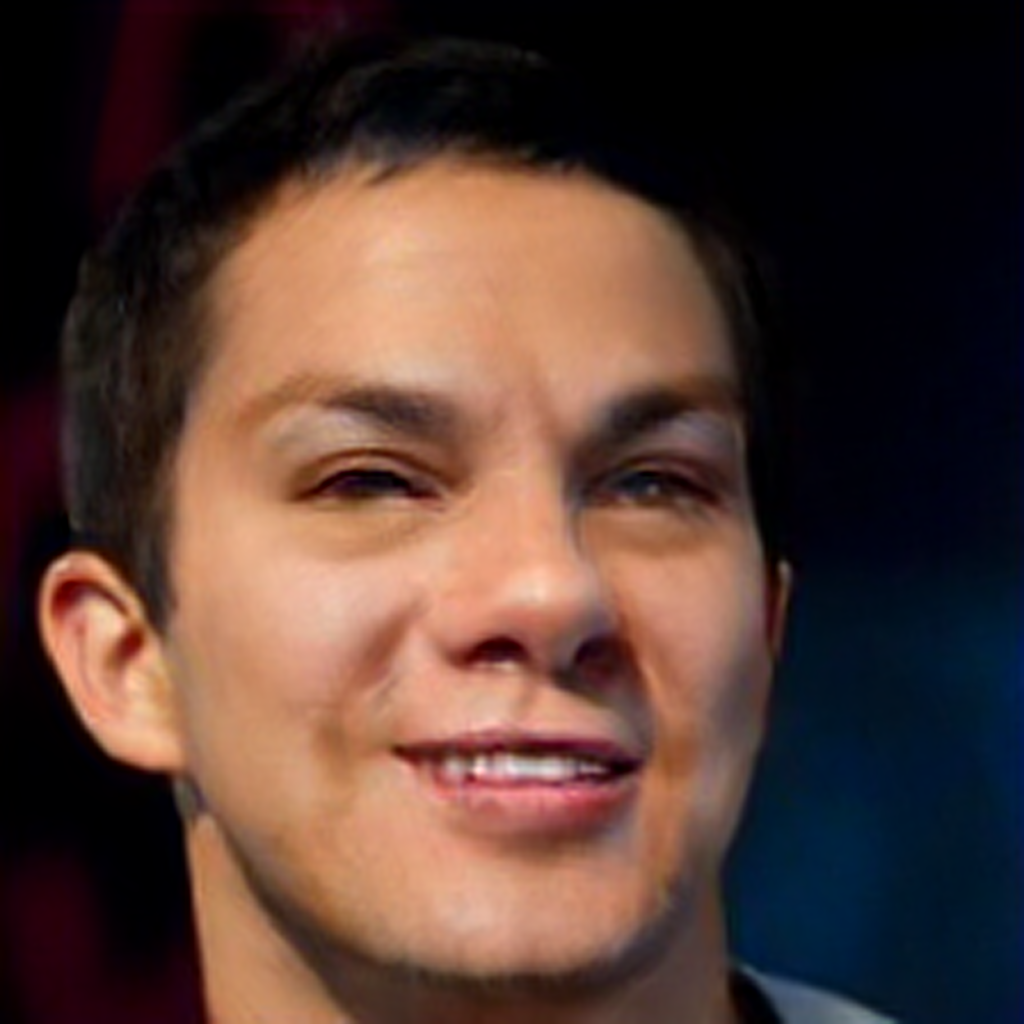} & \includegraphics[align=c,width=.12\textwidth]{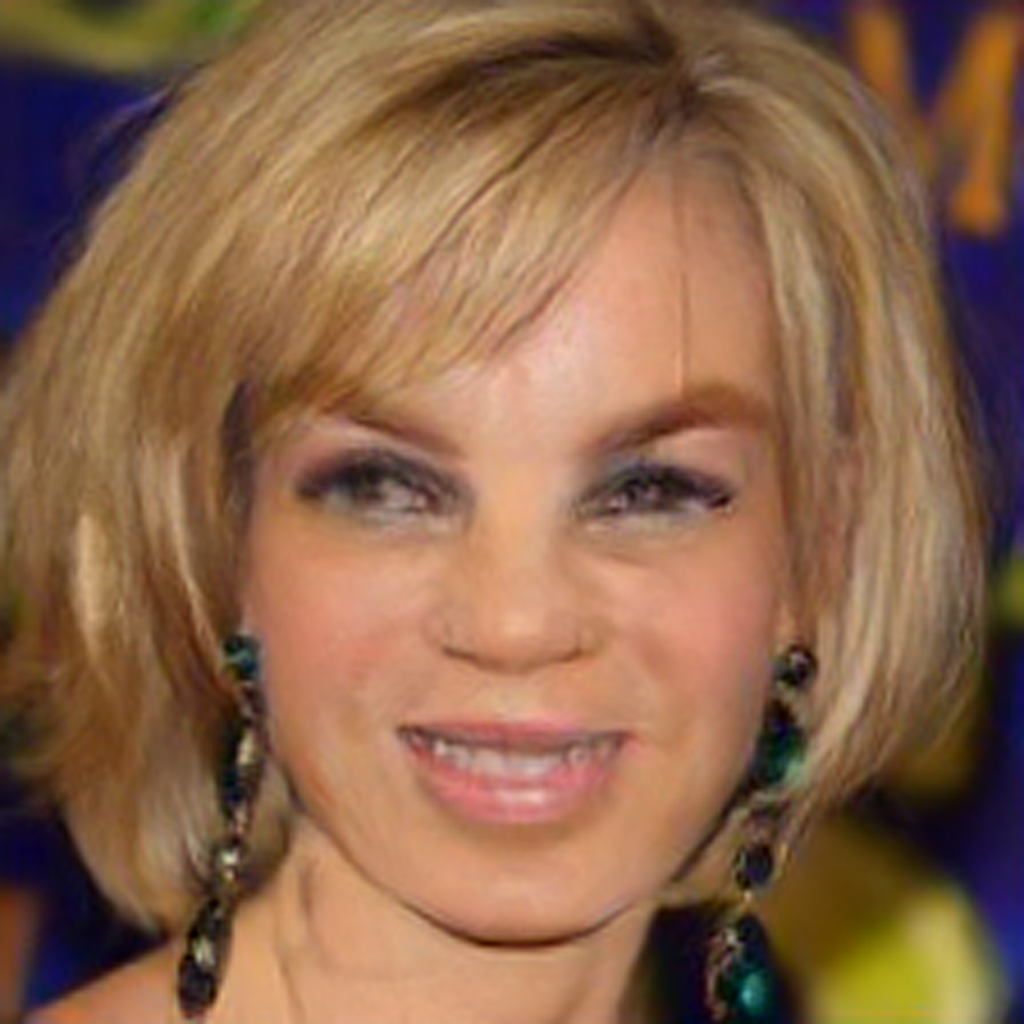} \vspace{.2em}\\
         AnonyGAN & \includegraphics[align=c,width=.12\textwidth]{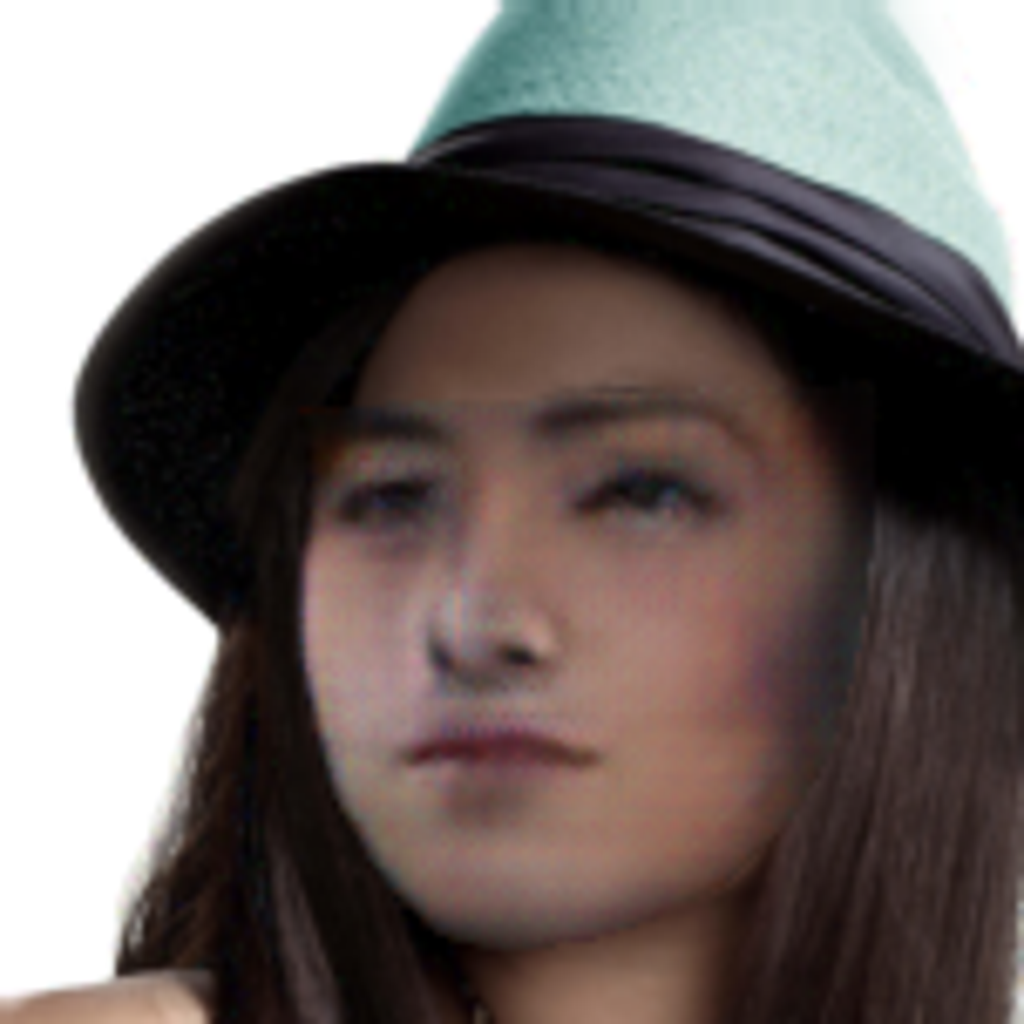} & \includegraphics[align=c,width=.12\textwidth]{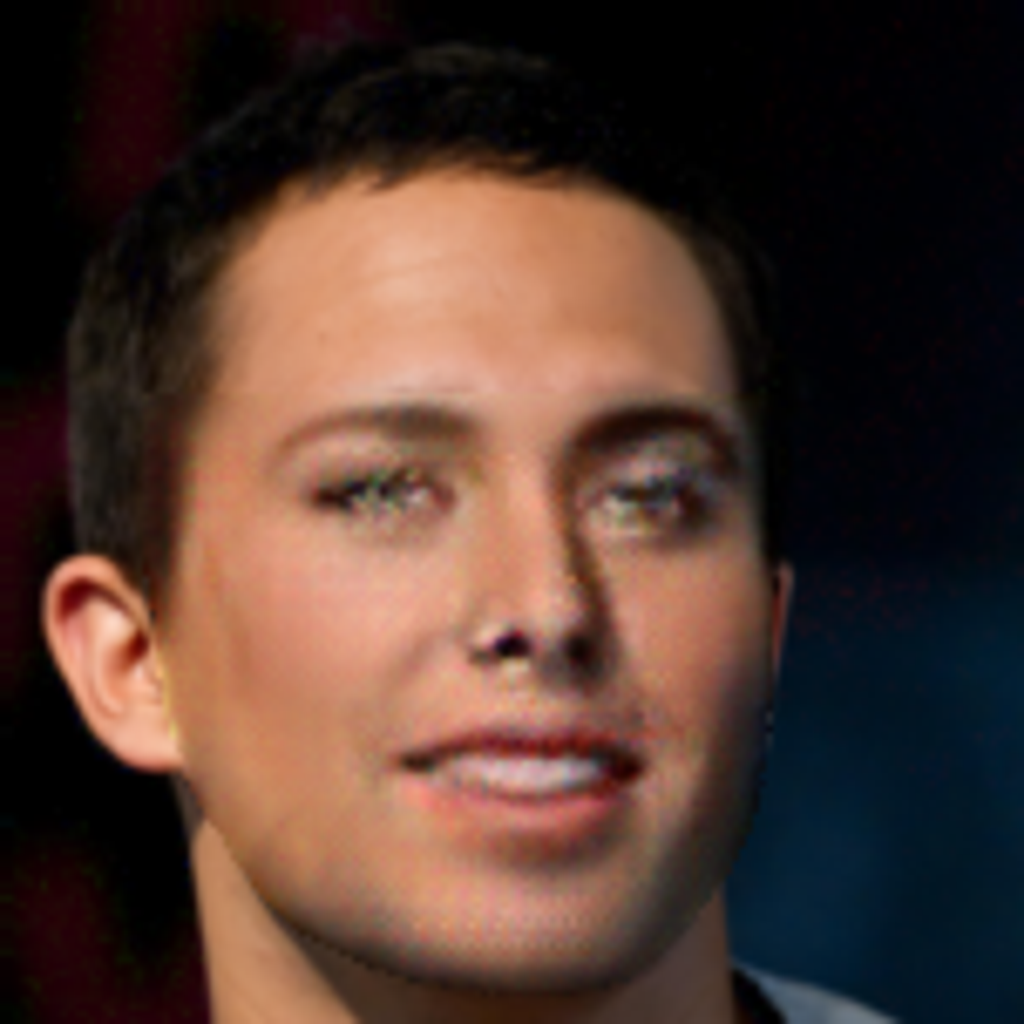} & \includegraphics[align=c,width=.12\textwidth]{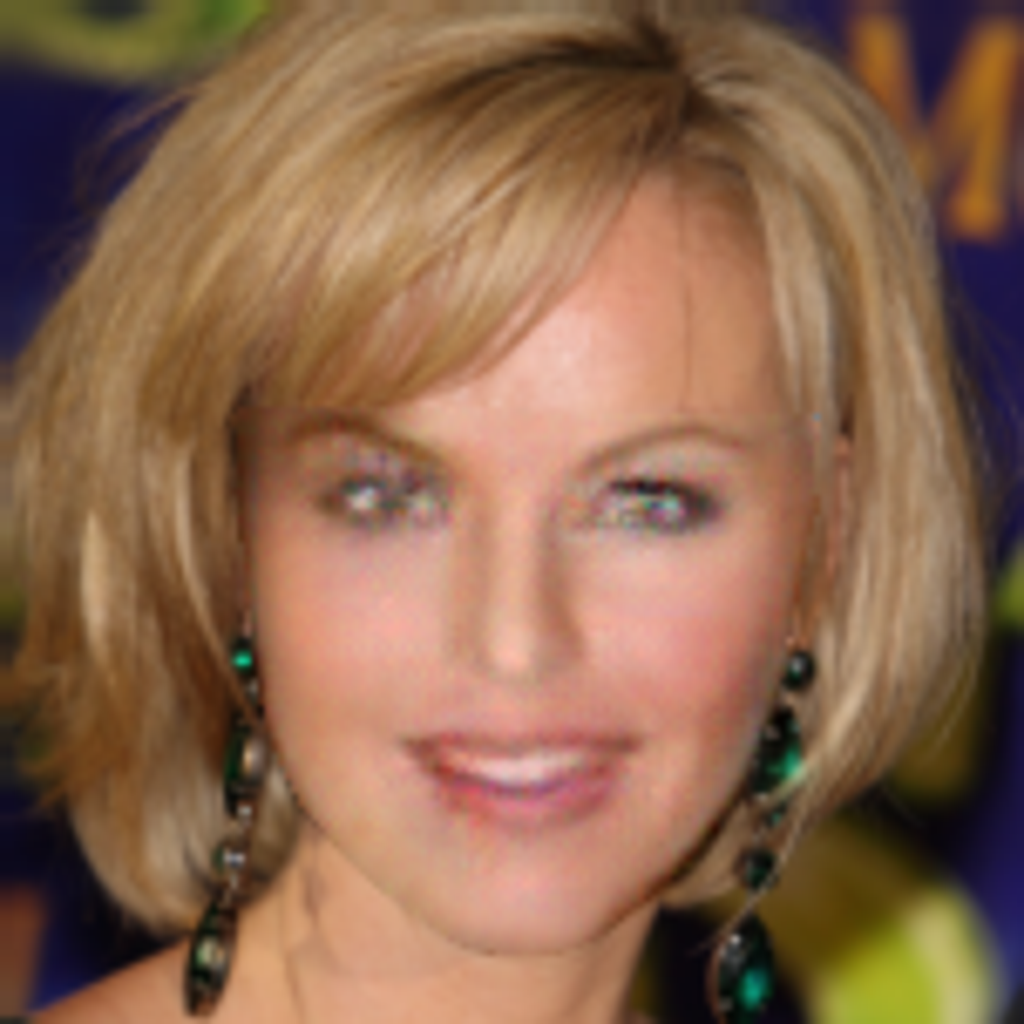} \vspace{.2em}\\
         CIAGAN & \includegraphics[align=c,width=.12\textwidth]{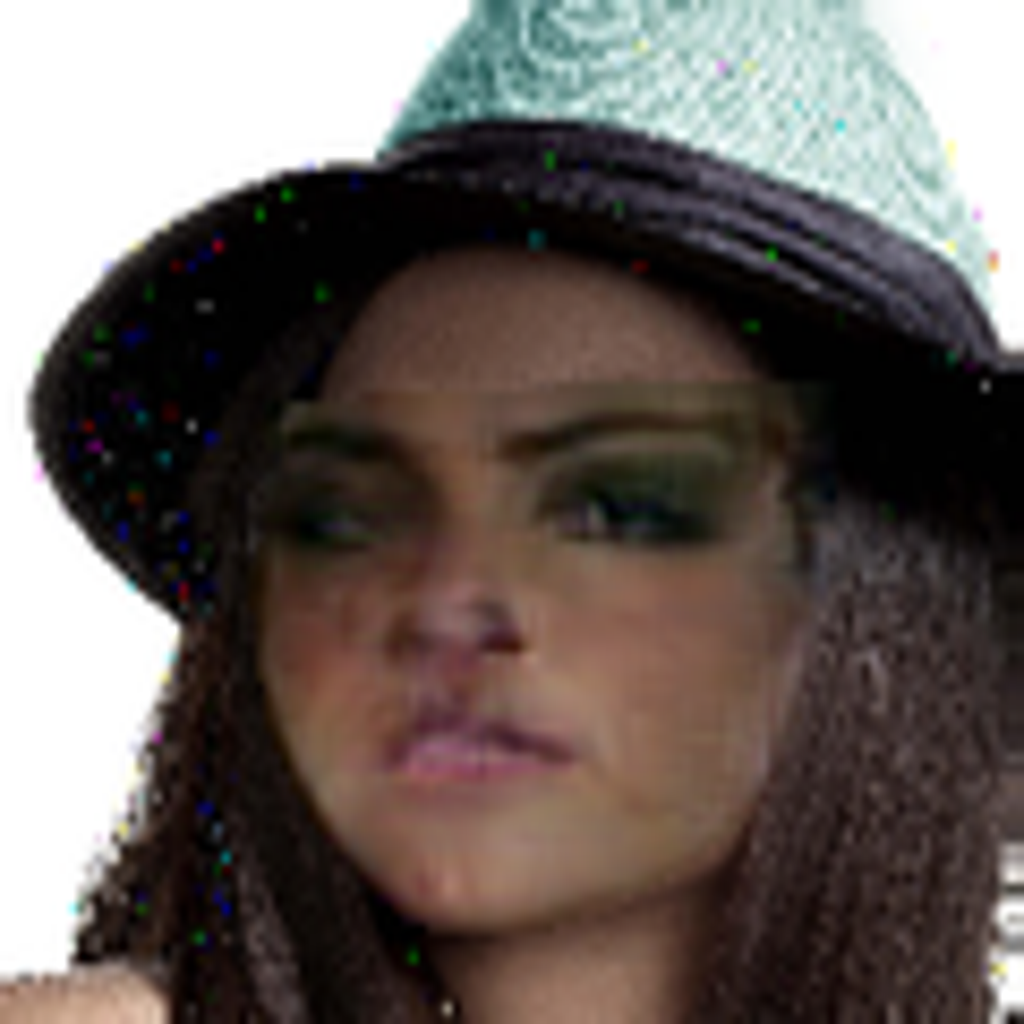} & \includegraphics[align=c,width=.12\textwidth]{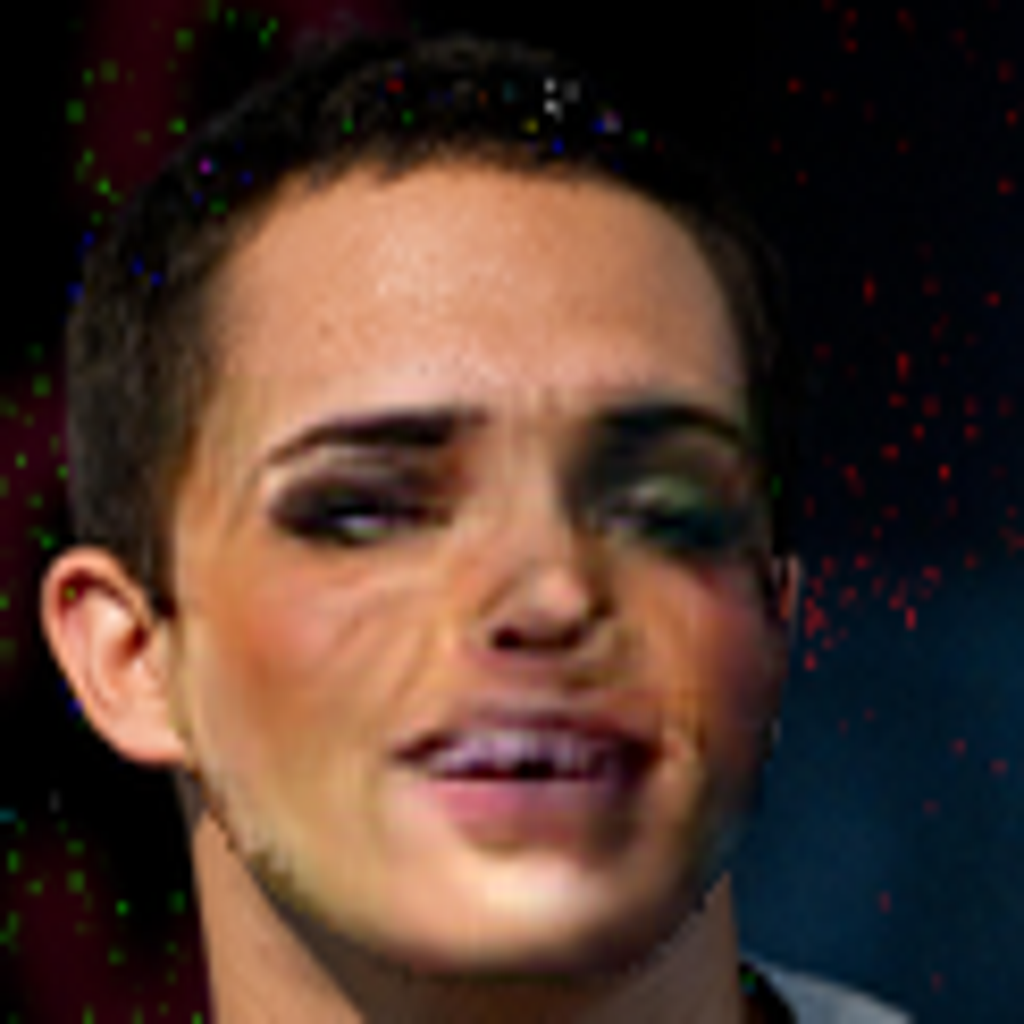} & \includegraphics[align=c,width=.12\textwidth]{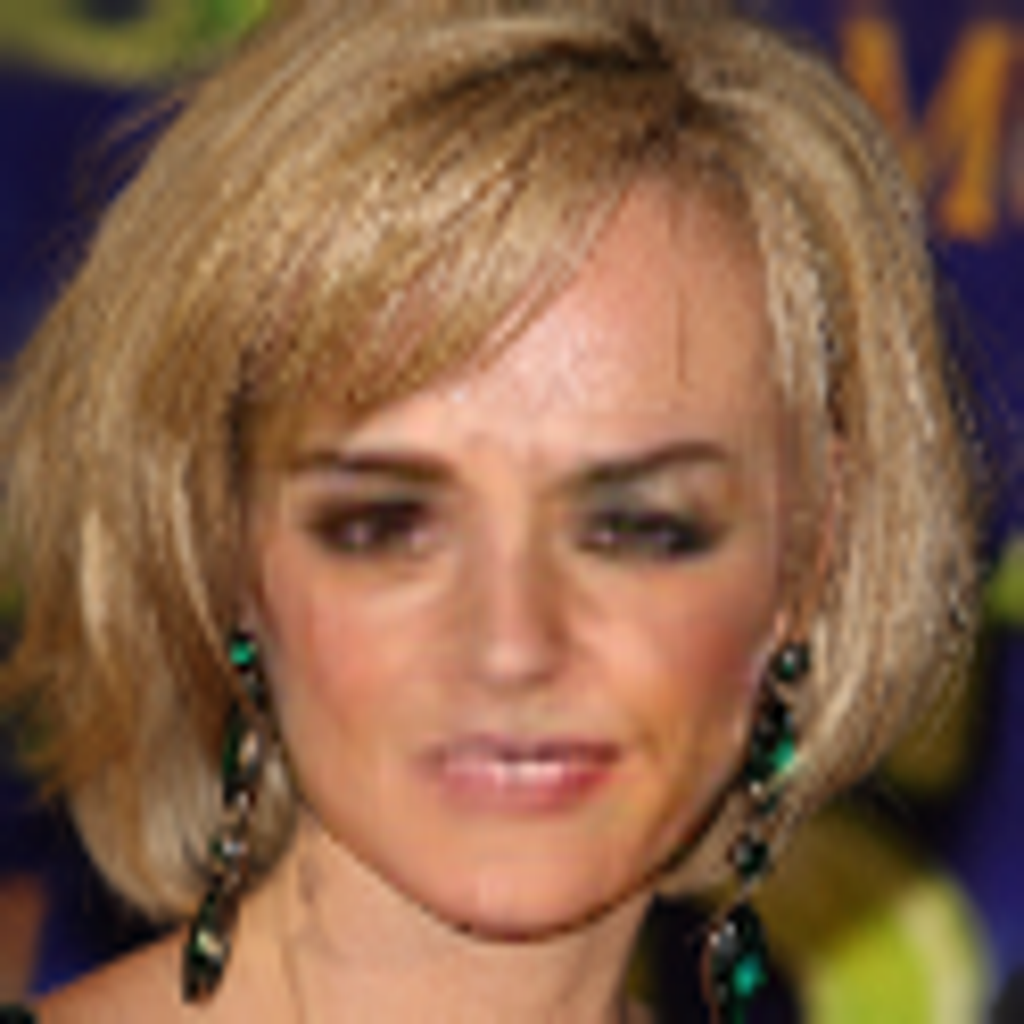}\vspace{.2em} \\
         DeepPrivacy & \includegraphics[align=c,width=.12\textwidth]{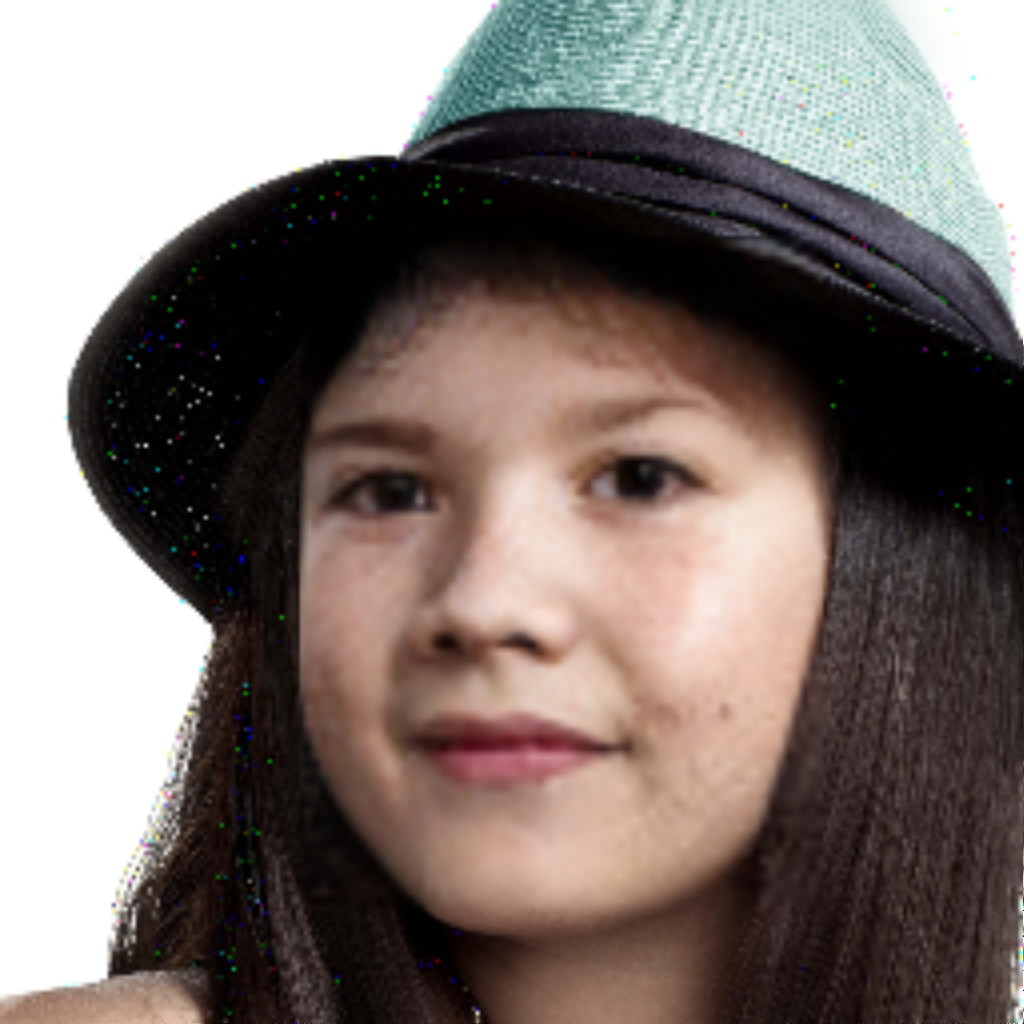} & \includegraphics[align=c,width=.12\textwidth]{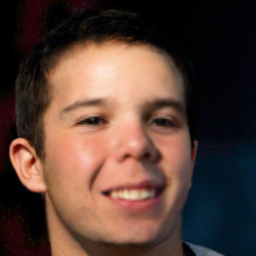} & \includegraphics[align=c,width=.12\textwidth]{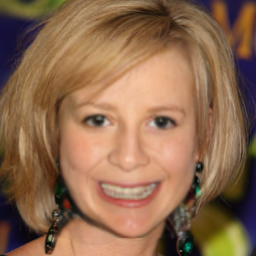} \vspace{.2em} \\
         DeepPrivacy2 & \includegraphics[align=c,width=.12\textwidth]{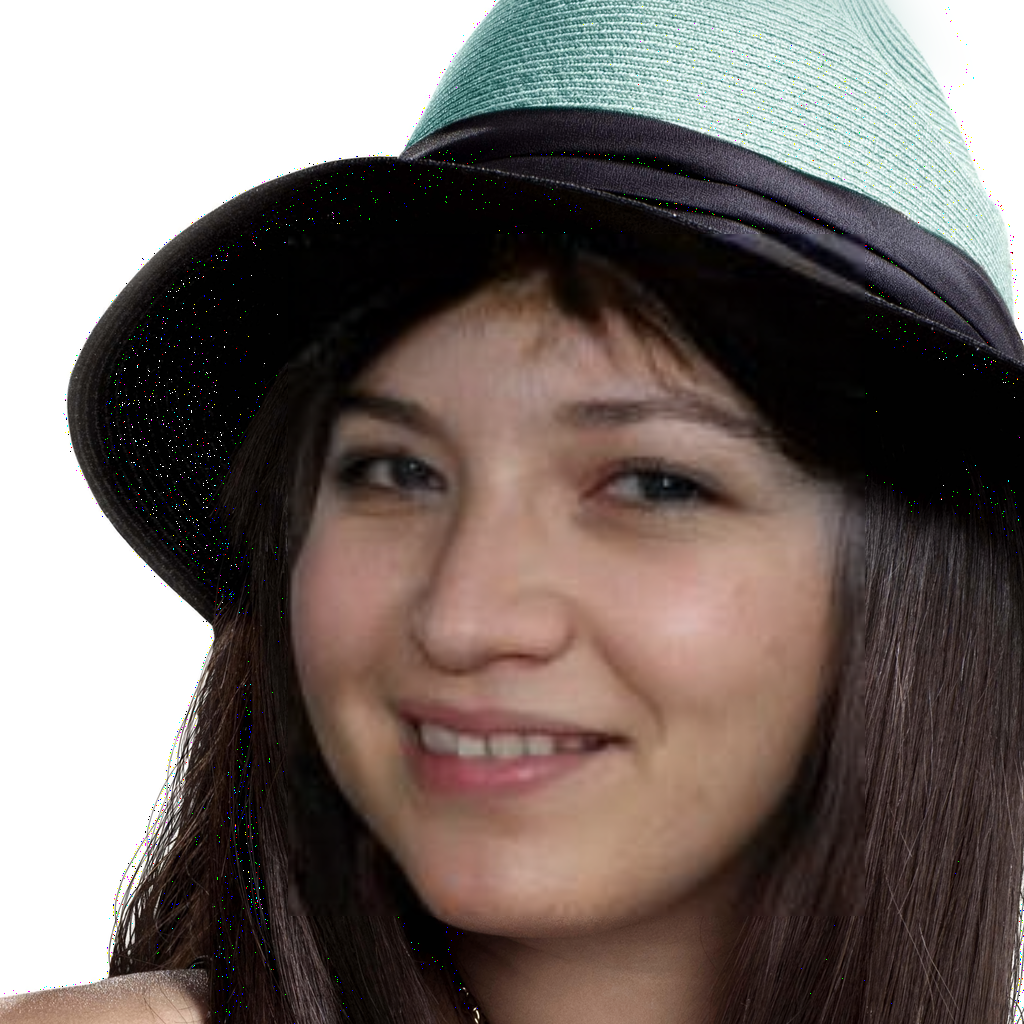} & \includegraphics[align=c,width=.12\textwidth]{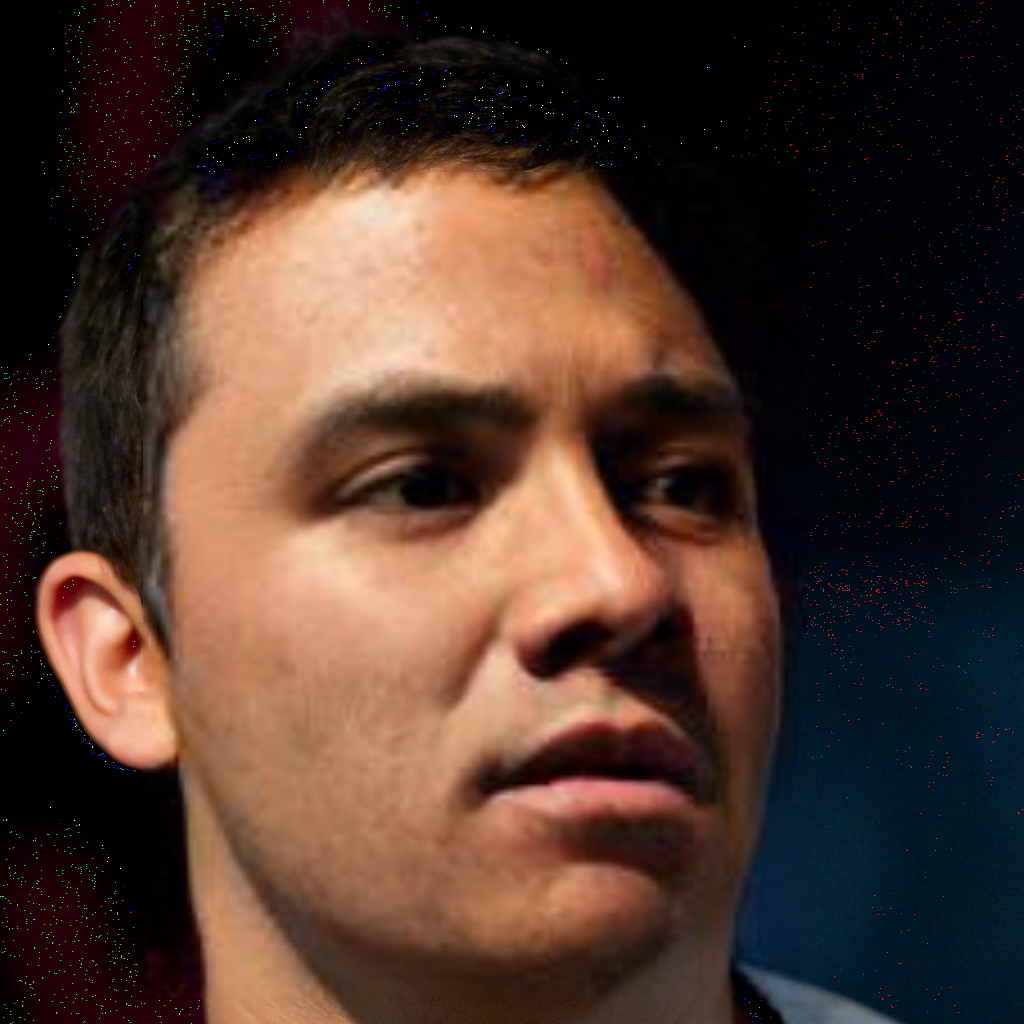} & \includegraphics[align=c,width=.12\textwidth]{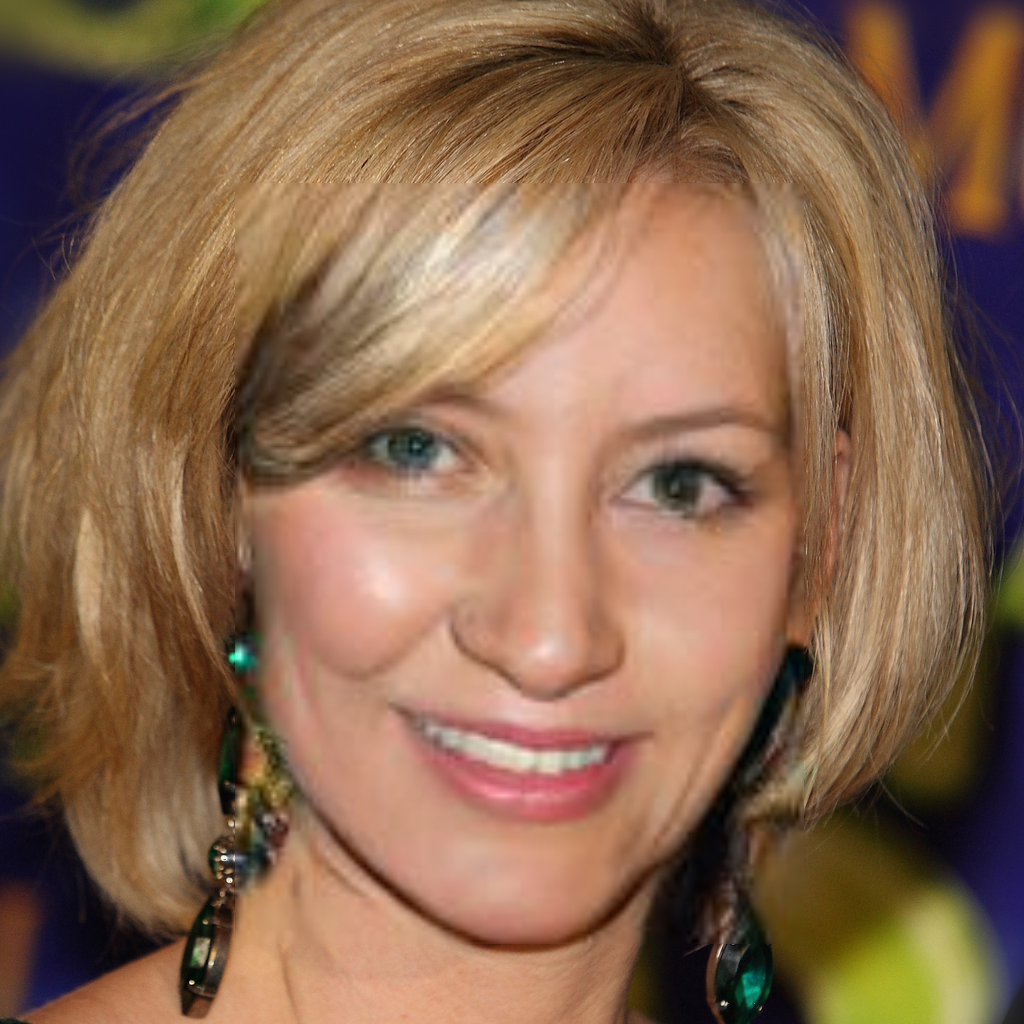}\vspace{.2em} \\
         FALCO & \includegraphics[align=c,width=.12\textwidth]{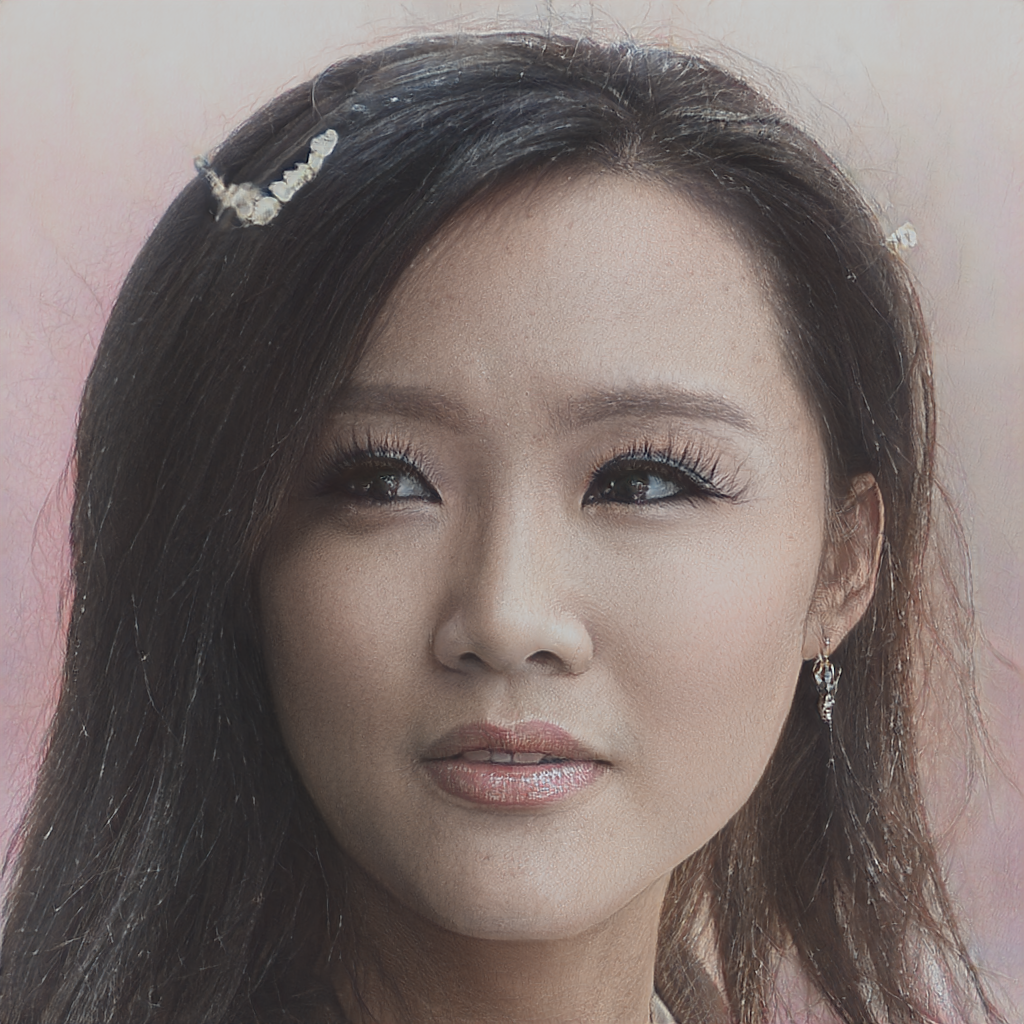} & \includegraphics[align=c,width=.12\textwidth]{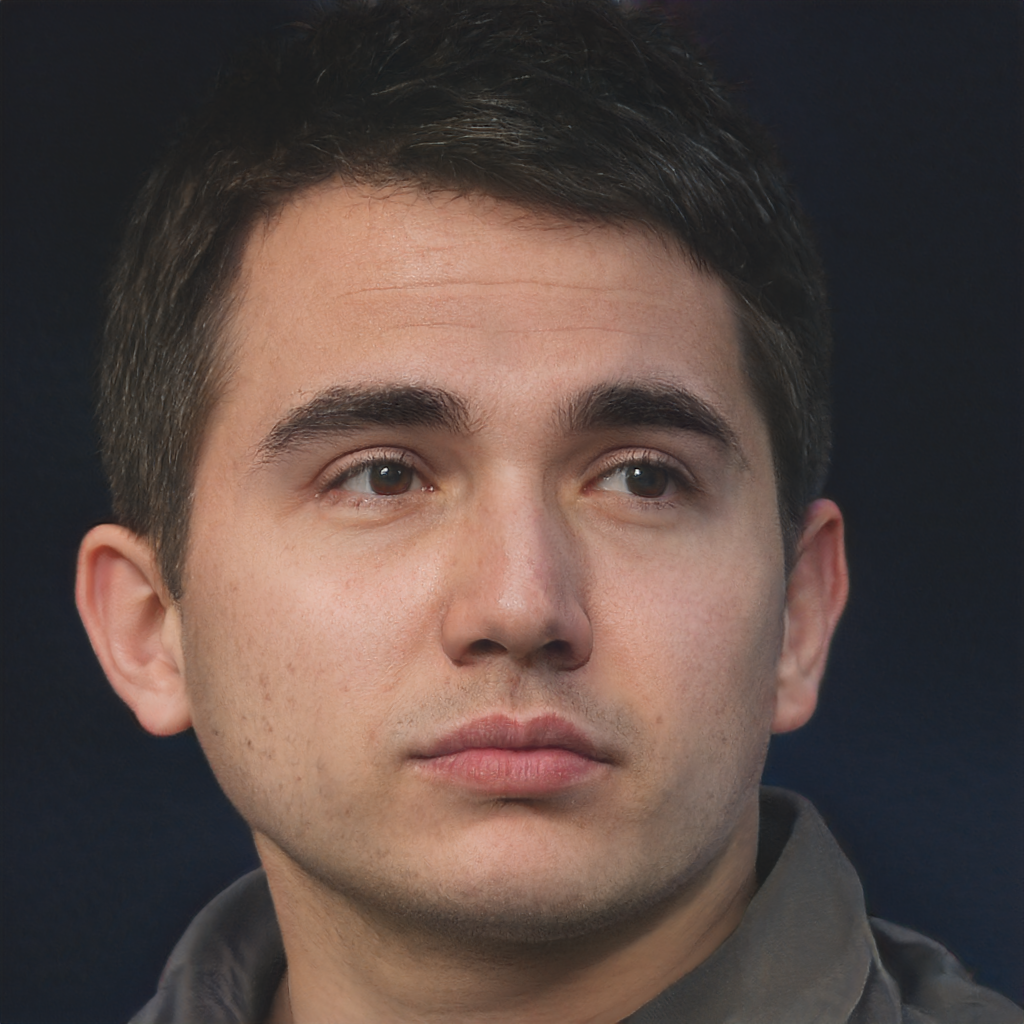} & \includegraphics[align=c,width=.12\textwidth]{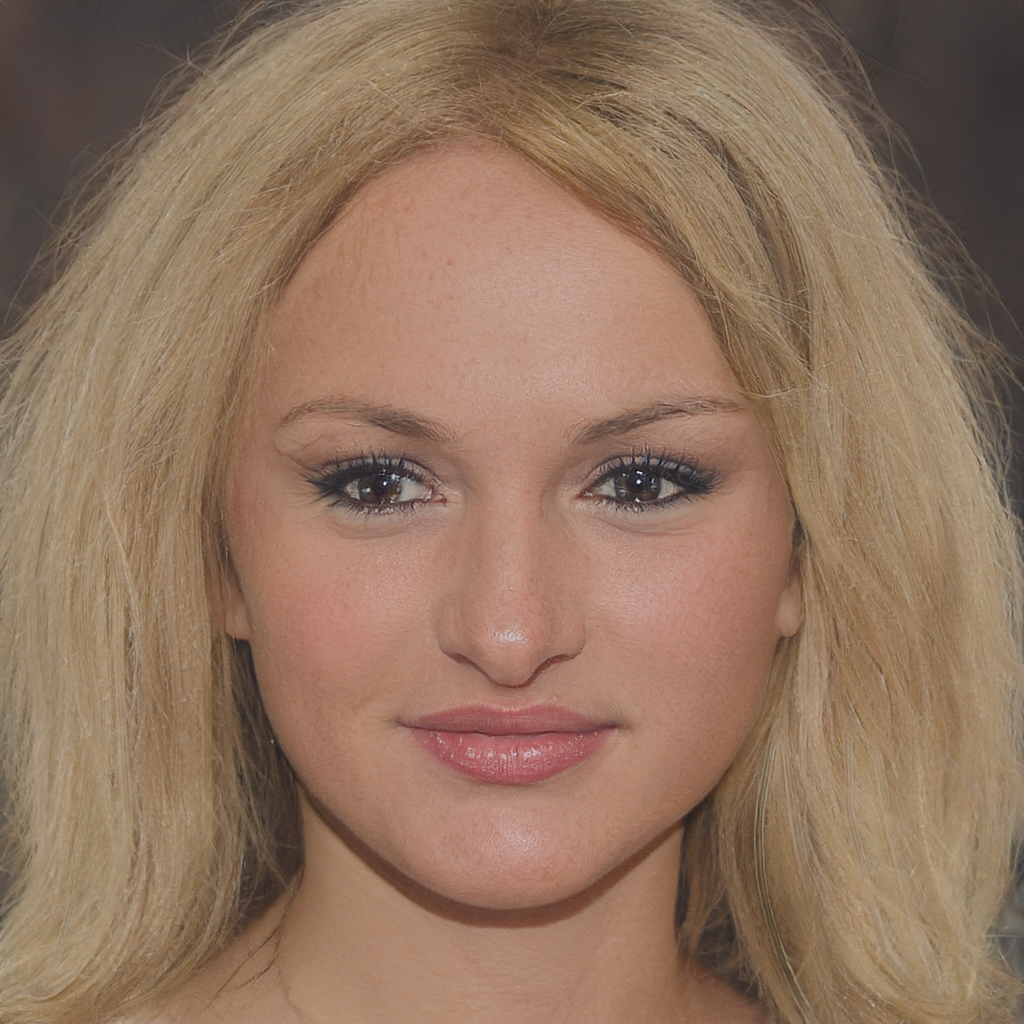} \vspace{.2em}\\
         G$^2$Face & \includegraphics[align=c,width=.12\textwidth]{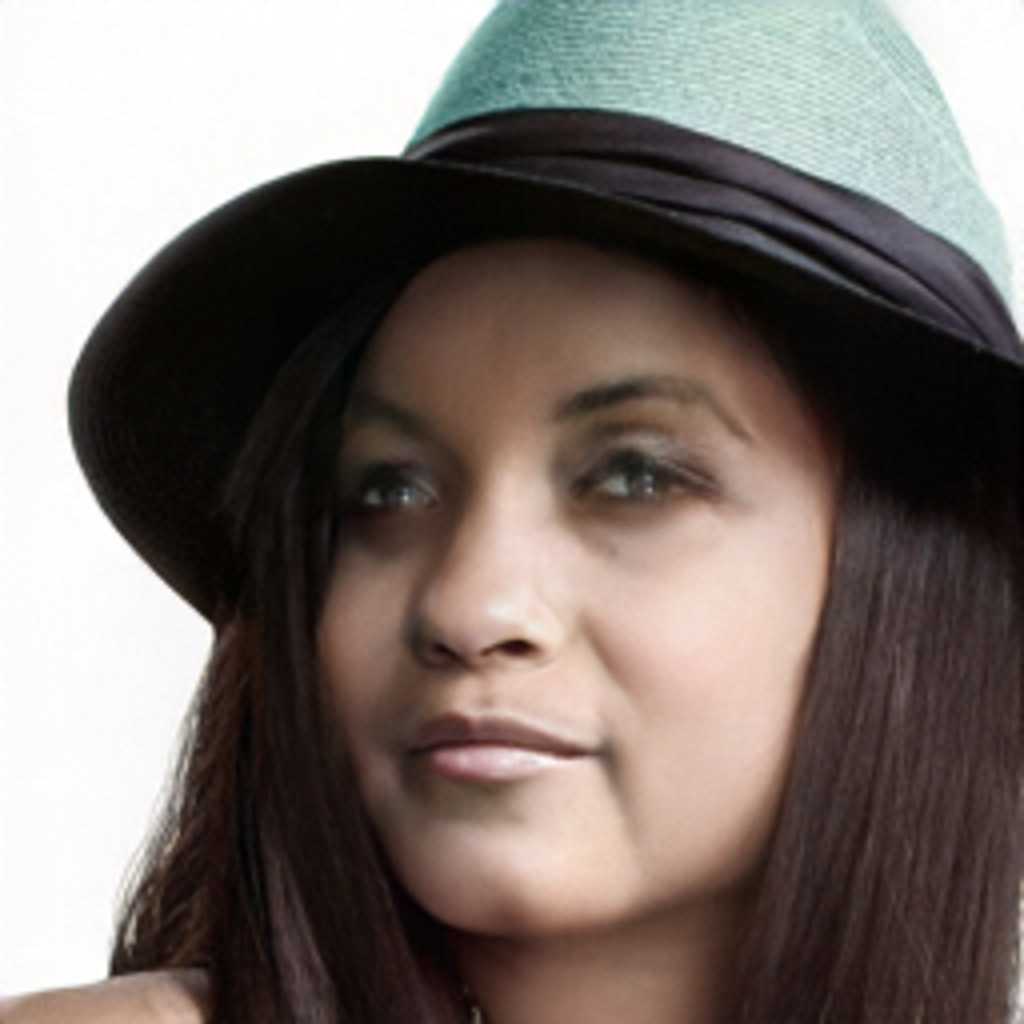} & \includegraphics[align=c,width=.12\textwidth]{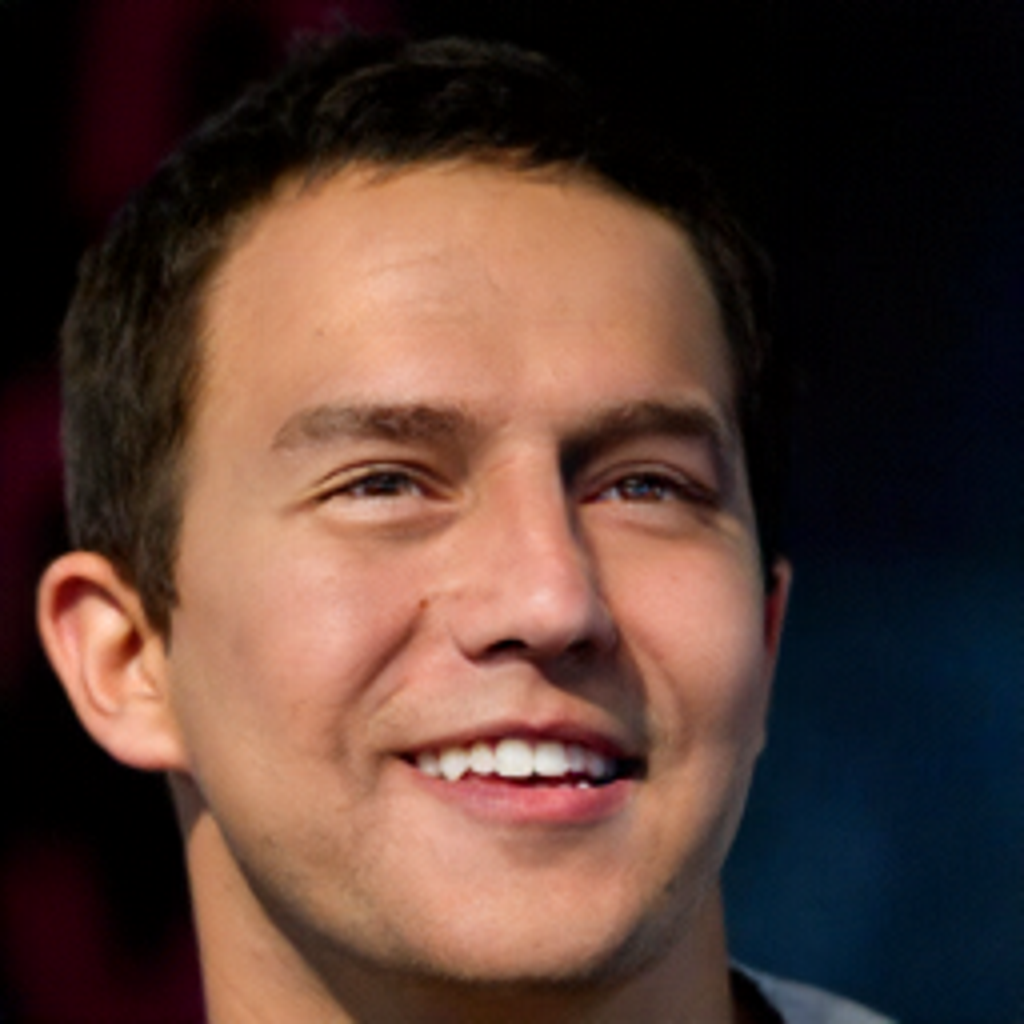} & \includegraphics[align=c,width=.12\textwidth]{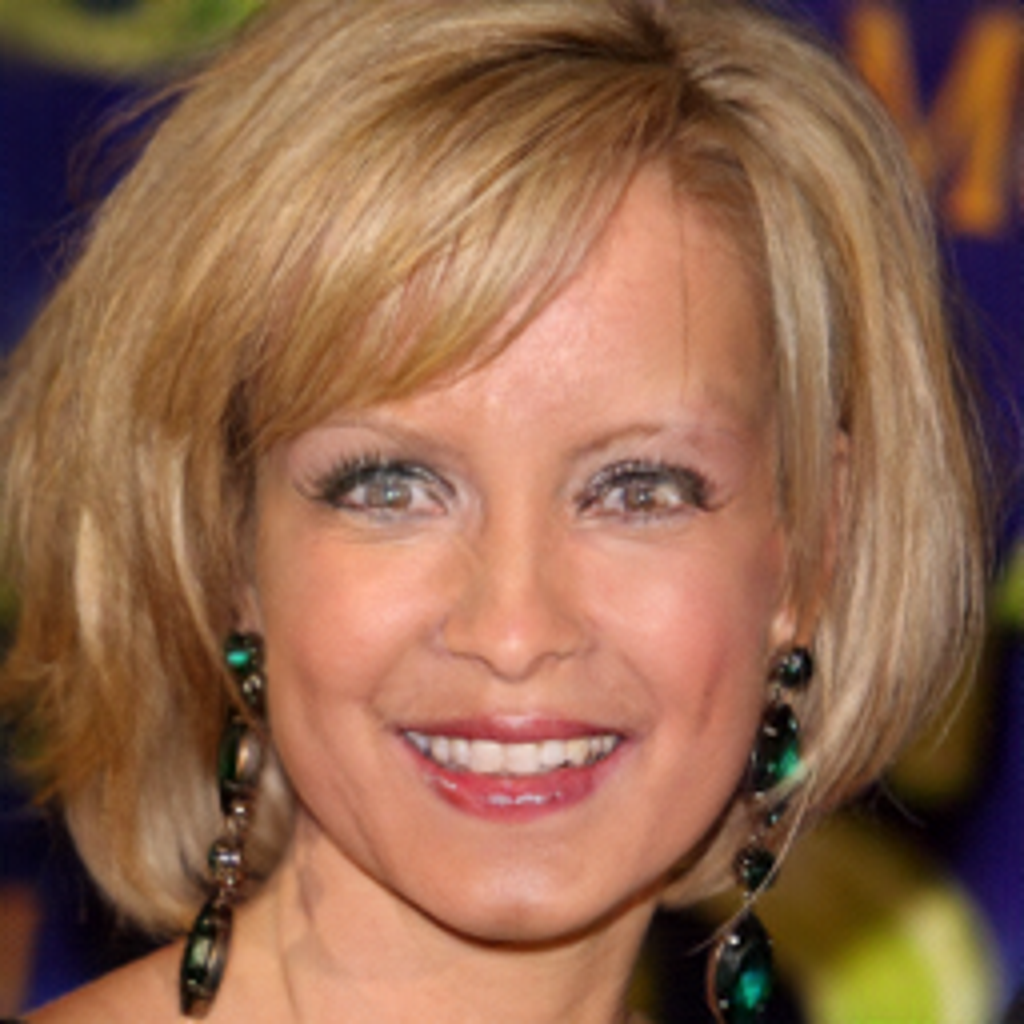} \vspace{.2em}\\
         RiDDLE & \includegraphics[align=c,width=.12\textwidth]{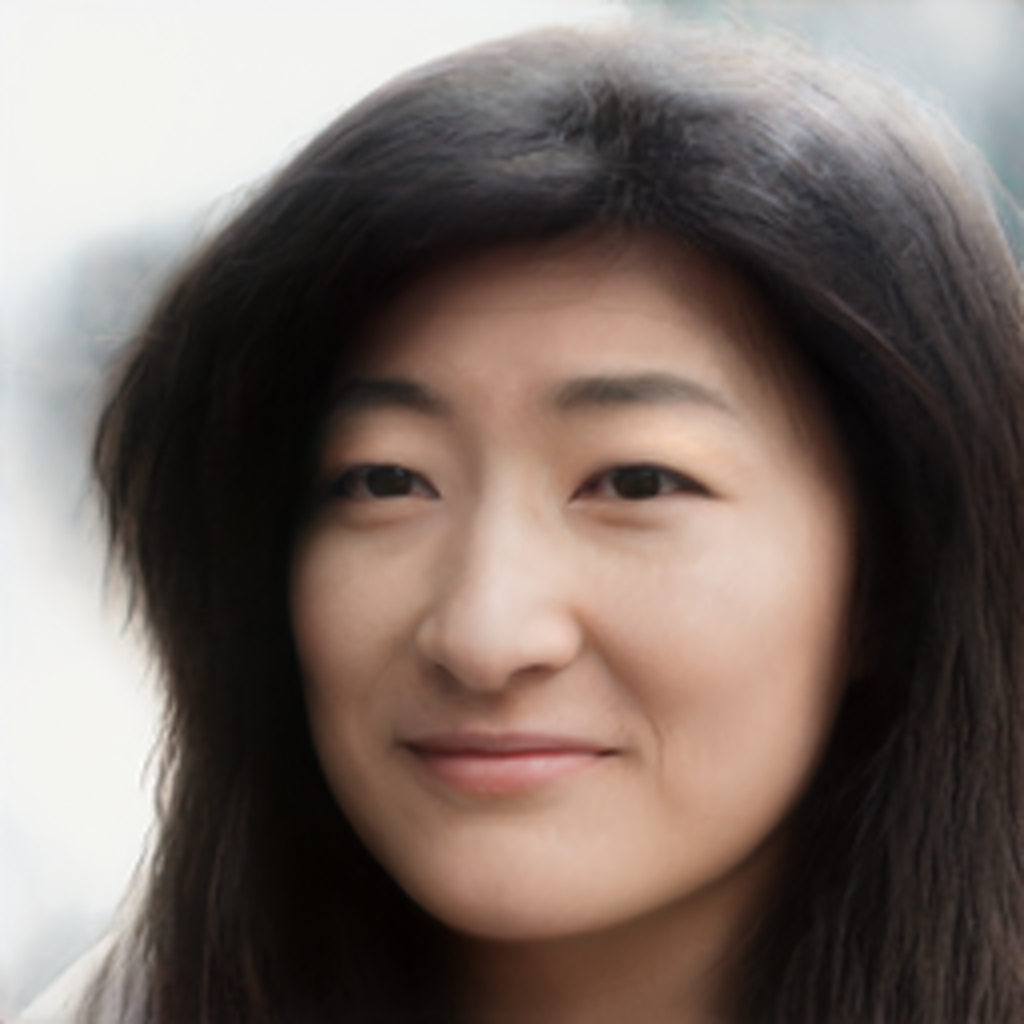} & \includegraphics[align=c,width=.12\textwidth]{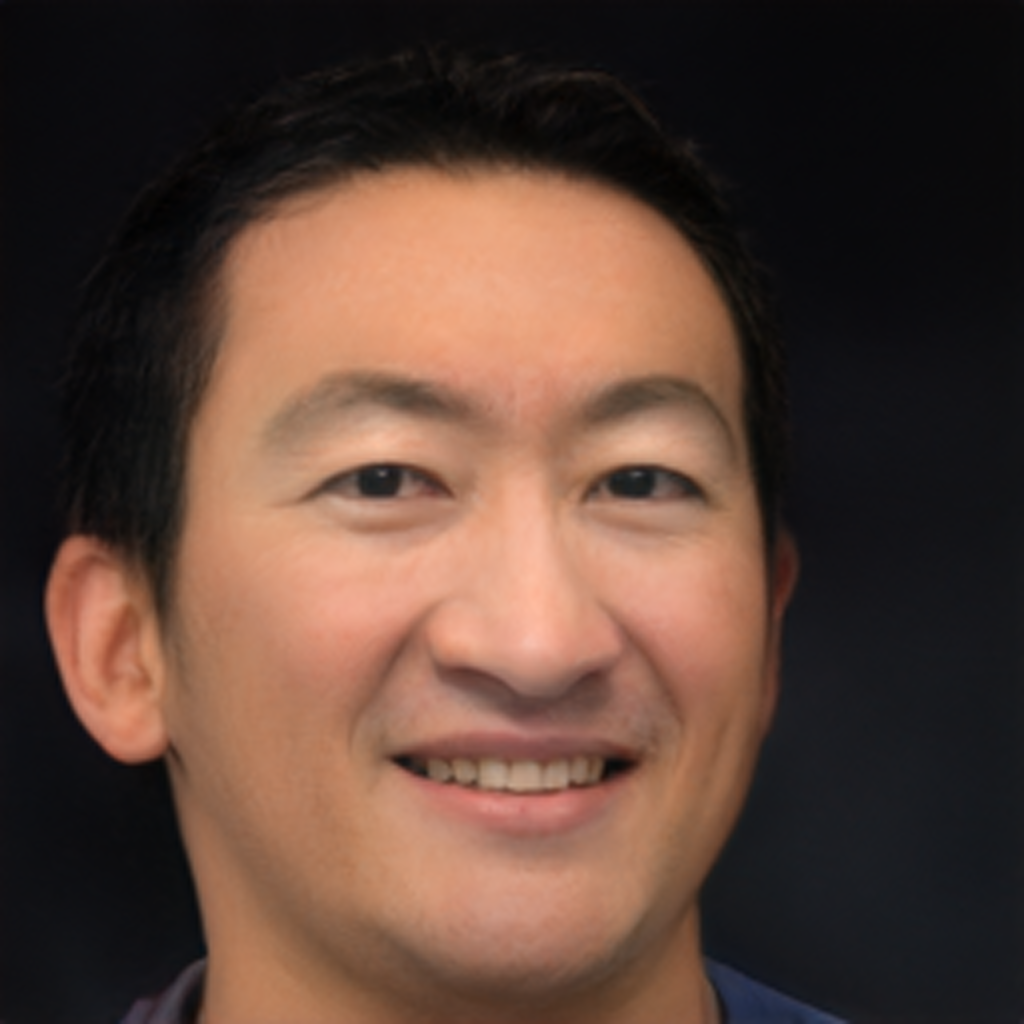} & \includegraphics[align=c,width=.12\textwidth]{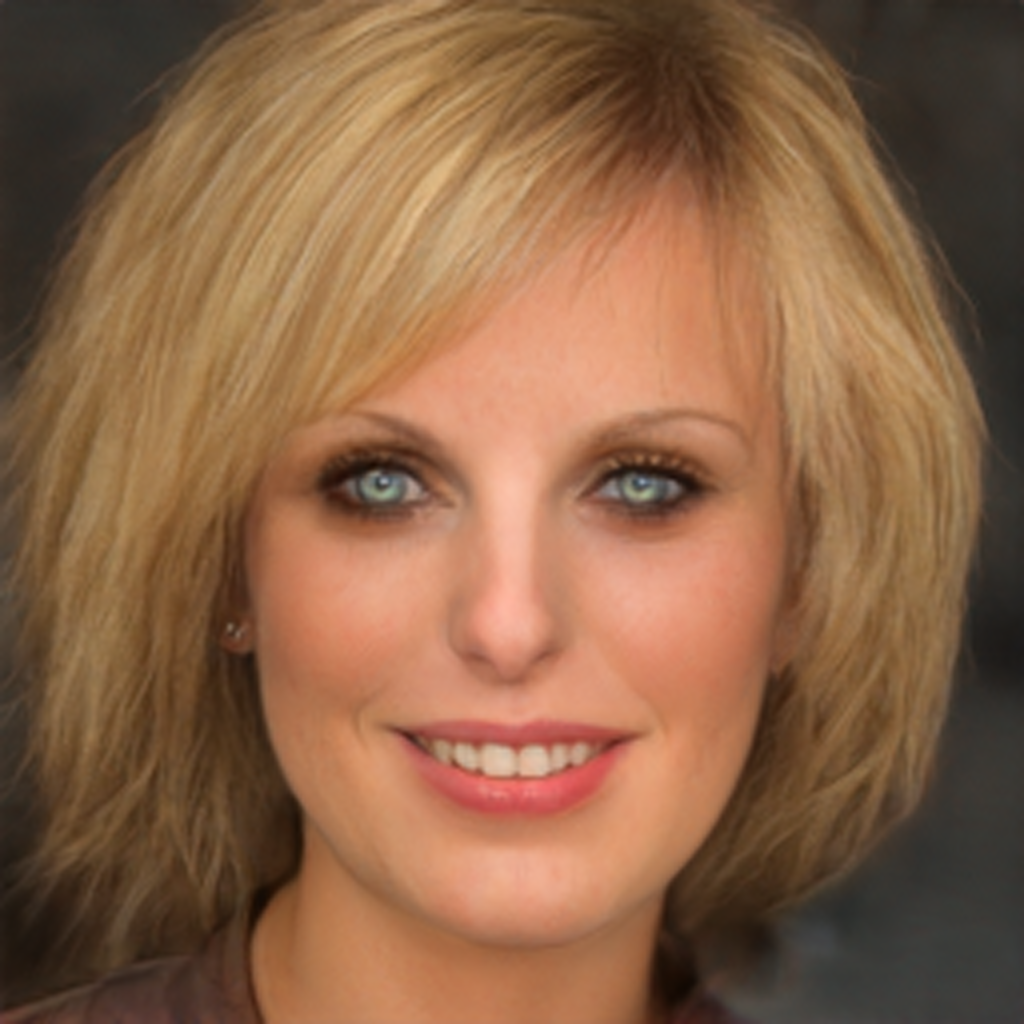} \vspace{.2em}\\
         Ours & \includegraphics[align=c,width=.12\textwidth]{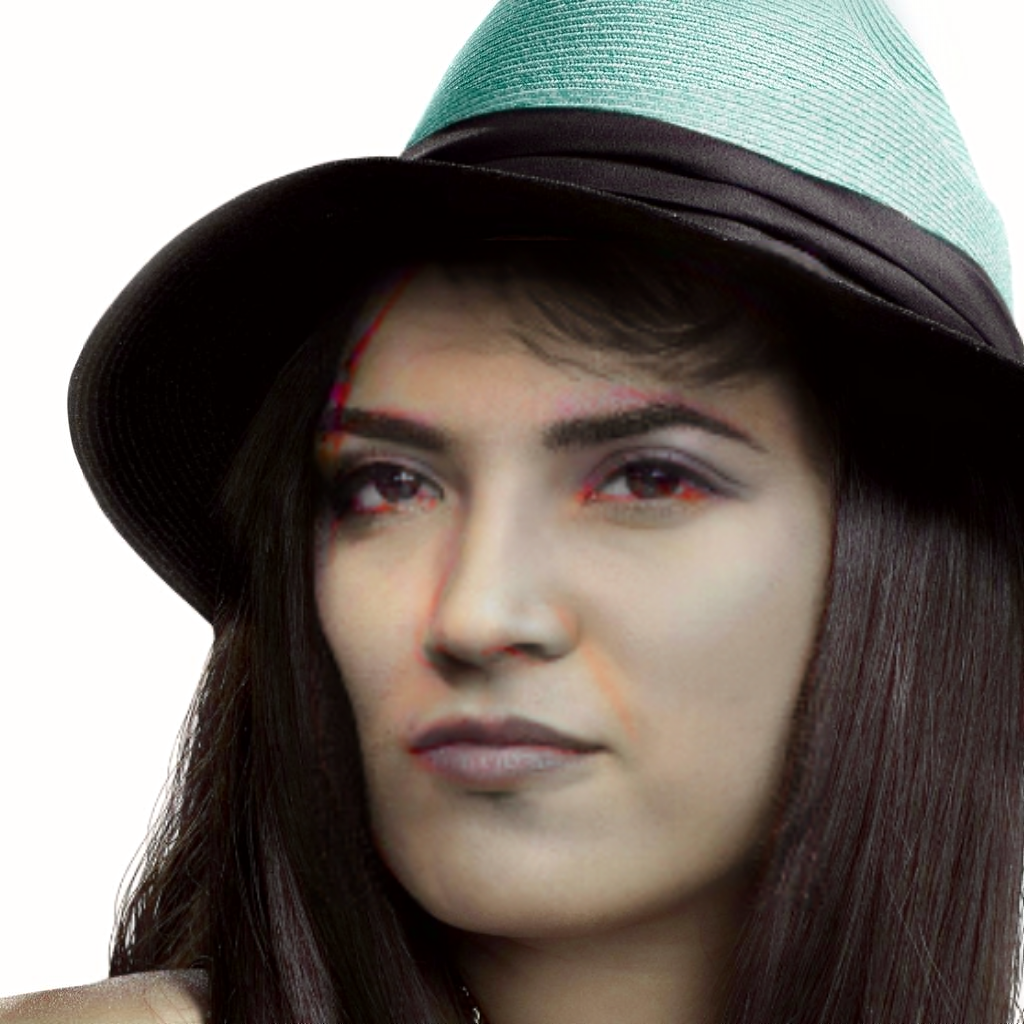} & \includegraphics[align=c,width=.12\textwidth]{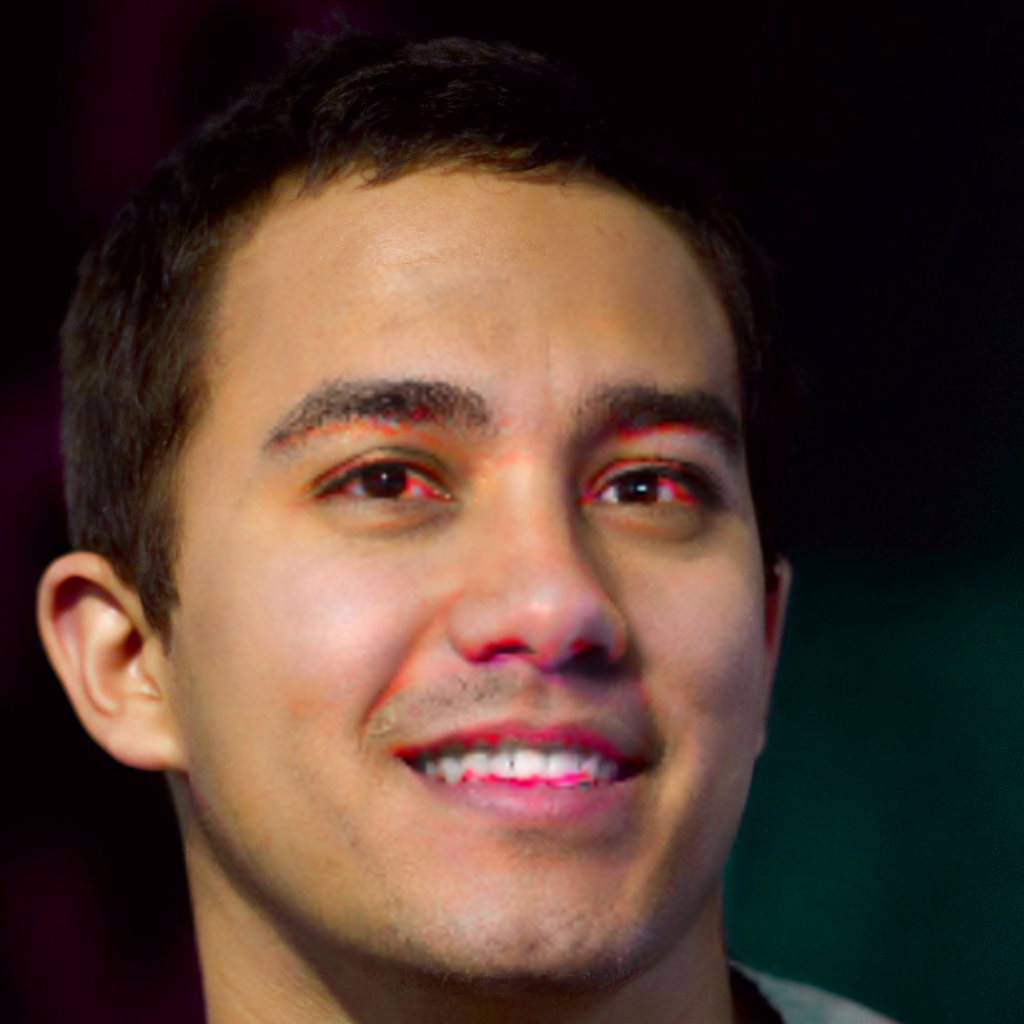} & \includegraphics[align=c,width=.12\textwidth]{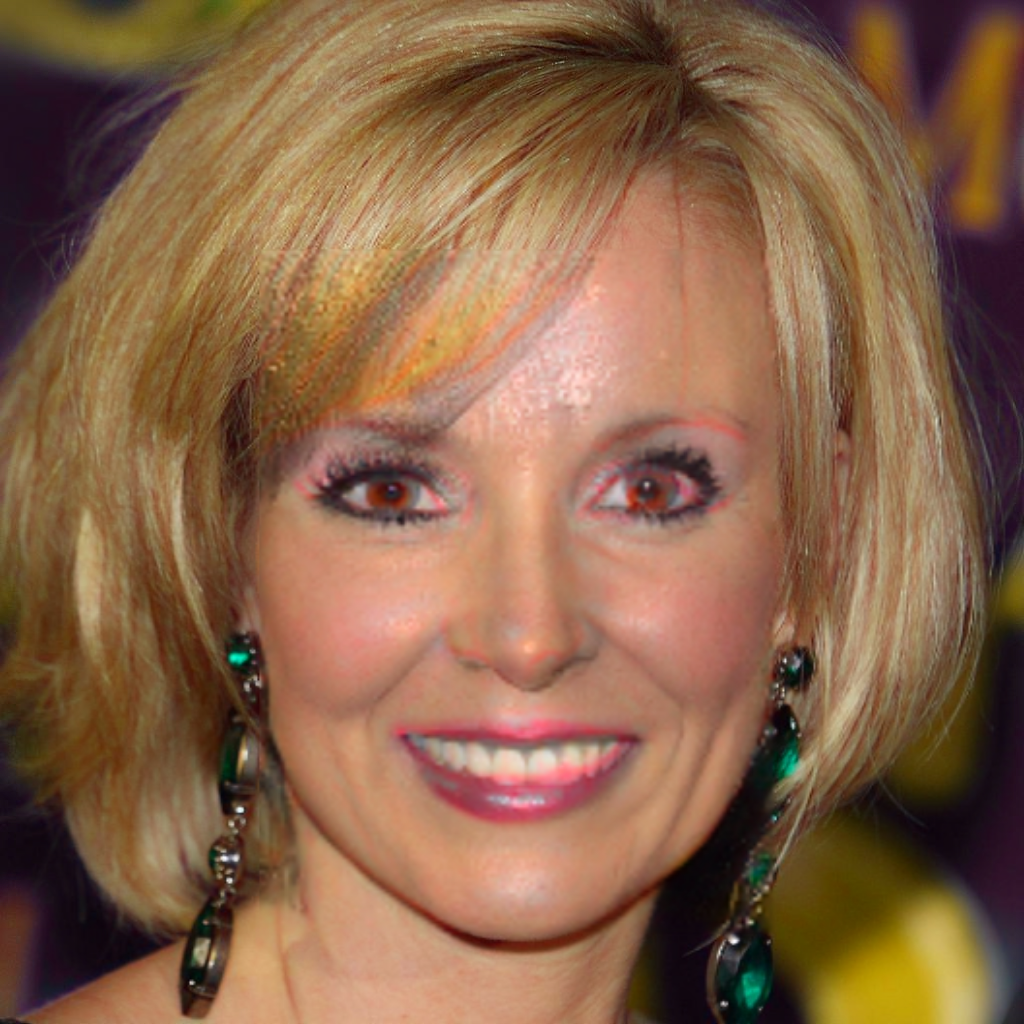}
    \end{tabular}
    \caption{Anonymization results on CelebA-HQ.}
    \label{fig:qualitative-results}
\end{figure}

\begin{table*}[ht!]
    \centering
    \caption{Comparison of common anonymization performance metrics across state-of-the-art methods. We report visual quality (FID$_{\textrm{ImageNet}}$, FID$_{\textrm{CLIP}}$), face detection rate, identity similarity using FaceNet embeddings (VGGFace2 and CASIA-WebFace), pose preservation error (normalized L$_1$ distance), and average inference time per image. $\downarrow$ indicates lower is better, $\uparrow$ indicates higher is better. Best and second-best results are highlighted in blue and red respectively.}
    \label{tab:common-metrics}
    \resizebox{\textwidth}{!}{\begin{tabular}{lcccccccc}
    \toprule
        \multirow{2}{*}{Method} & \multirow{2}{*}{Year} & \multirow{2}{*}{FID$_{\textrm{ImageNet}}$ ($\downarrow$)} & \multirow{2}{*}{FID$_{\textrm{CLIP}}$ ($\downarrow$)} & \multirow{2}{*}{Face detection ($\uparrow$)} & \multicolumn{2}{c}{Face re-identification ($\downarrow$)} & \multirow{2}{*}{Pose estimation ($\downarrow$)} & \multirow{2}{*}{Inference time ($\downarrow$)} \\ 
        & & & & & VGGFace2 & CASIA-WebFace \\
        \midrule
        AIDPro \cite{wang2025beyond}                    & 2025 & 26.63 & 14.48 & 98.79 & 0.1532 $\pm$ 0.1591 & 0.2140 $\pm$ 0.1470 & 0.0500 & \best{0.02} \\
        AnonyGAN \cite{dall2022graph}                   & 2022 & 60.23 & 19.34 & 96.79 & 0.3892 $\pm$ 0.1546 & 0.3768 $\pm$ 0.1408 & 0.0339 & 0.10 \\
        CIAGAN \cite{maximov2020ciagan}                 & 2020 & \best{1.78} & 34.00 & 95.25 & 0.1796 $\pm$ 0.1658 & 0.2172 $\pm$ 0.1432 & 0.0791 & 0.08 \\
        DeepPrivacy \cite{hukkelaas2019deepprivacy}     & 2019 & 32.50 & 21.10 & 97.24 & 0.1917 $\pm$ 0.1748 & 0.2170 $\pm$ 0.1514 & 0.0520 & 0.12 \\
        DeepPrivacy2 \cite{hukkelaas2022deepprivacy2}   & 2022 & 15.08 & 11.69 & 95.84 & 0.1738 $\pm$ 0.1681 & 0.2063 $\pm$ 0.1507 & 0.0946 & 0.28 \\
        FALCO \cite{barattin2023attribute}              & 2023 & 26.59 & 14.46 & \secondbest{99.74} & 0.4822 $\pm$ 0.1439 & 0.4996 $\pm$ 0.1268 & 0.0598 & 0.09 \\ 
        $\mathrm{G^2}$Face~\cite{yang2024g}             & 2024 & 16.32 & \best{5.77} & 98.38 & \secondbest{0.1430 $\pm$ 0.1632} & \secondbest{0.1916 $\pm$ 0.1485} & \best{0.0276} & \secondbest{0.03} \\ 
        RiDDLE~\cite{li2023riddle}                      & 2023 & 52.79 & 33.63 & \best{99.86} & \best{0.0326 $\pm$ 0.1683} & \best{0.0941 $\pm$ 0.1513} & 0.0446 & \secondbest{0.03} \\ \midrule
        Our          & 2025 & \secondbest{10.54} & \secondbest{8.27} & 98.04 & 0.2125 $\pm$ 0.1599 & 0.2184 $\pm$ 0.1420 & \secondbest{0.0334} & 0.12 \\
        \bottomrule 
    \end{tabular}}
\end{table*}
Table~\ref{tab:common-metrics} reports the results on the common evaluation metrics, including image quality (FID$_{\textrm{ImageNet}}$ and FID$_{\textrm{CLIP}}$), face detection rate, face re-identification similarity, and pose preservation. Our method achieves the second-best FID$_{\textrm{ImageNet}}$ and FID$_{\textrm{CLIP}}$ scores, indicating that the generated images are perceptually close to the distribution of real faces while avoiding overfitting to the ImageNet feature space. In particular, our CLIP-based FID is substantially lower than DeepPrivacy2, FALCO, and AIDPro, demonstrating the semantic consistency of the generated content. Although RiDDLE and G$^2$Face achieve slightly better re-identification suppression, our method provides a strong balance between privacy and visual fidelity, outperforming AnonyGAN and DeepPrivacy in both FID and identity obfuscation. Pose preservation is also competitive, with our normalized landmark error being the second lowest across all methods, suggesting that our approach reliably maintains the geometric structure of the input faces. While our inference time is not the fastest, it remains practical for real-time or batch anonymization scenarios, with only a marginal increase compared to the fastest baselines. A visual comparison of anonymized samples is shown in Figure~\ref{fig:qualitative-results}, where our method exhibits stronger preservation of expression, shading, and skin tone, with fewer visual artifacts than the competing methods.

Although RiDDLE and G$^2$Face achieve slightly better identity suppression, our method strikes a stronger balance between privacy protection, photometric fidelity, and emotional realism, offering a more versatile solution for practical anonymization scenarios.

Table~\ref{tab:emotion-evaluation} focuses on the preservation of facial expressions by reporting both landmark error (on 51 expression-related points) and emotion classification accuracy. Our method attains the highest expression classification accuracy (75.00\%), outperforming AnonyGAN and G$^2$Face, which are the closest competitors. This result highlights the ability of our model to preserve subtle expression cues, which is essential for downstream tasks such as affective computing and human behavior analysis. The landmark error achieved by our method is also the second lowest overall, confirming that expression-related facial geometry is faithfully preserved.
\begin{table}
    \centering
    \caption{Evaluation of facial expression preservation using normalized landmark error (51 expression-related landmarks) and emotion classification accuracy. Lower values ($\downarrow$) indicate better landmark alignment, higher values ($\uparrow$) indicate better expression consistency. Best and second-best results are highlighted in blue and red respectively.}
    \label{tab:emotion-evaluation}
    \begin{tabular}{lcc}
    \toprule
         Method & Landmark error ($\downarrow$) & Facial expression acc. ($\uparrow$) \\ \midrule
         AIDPro \cite{wang2025beyond} & 0.0510 & 63.91 \\ 
         AnonyGAN \cite{dall2022graph} & 0.0321 & \secondbest{73.11} \\
         CIAGAN \cite{maximov2020ciagan} & 0.0801 & 56.55 \\
         DeepPrivacy \cite{hukkelaas2019deepprivacy} & 0.0513 & 53.29 \\
         DeepPrivacy2 \cite{hukkelaas2022deepprivacy2} & 0.0973 & 51.23 \\
         FALCO \cite{barattin2023attribute} & 0.0598 & 59.43 \\
         $\mathrm{G^2}$Face~\cite{yang2024g} & \best{0.0252} & 72.00 \\
         RiDDLE \cite{li2023riddle} & 0.0446 & 67.66 \\
         \midrule
         Our & \secondbest{0.0318} & \best{75.00} \\ \bottomrule
    \end{tabular}
\end{table}

Table~\ref{tab:lighting-comparison} presents the evaluation of lighting direction preservation using scale-invariant metrics. Our method achieves the lowest Si-L$_2$ and a near-best Si-MSE, outperforming all other baselines, including G$^2$Face, which is specifically designed to preserve geometric properties. This suggests that our approach not only maintains spatial structure but also captures shading consistency, which is crucial for photorealistic anonymization and for downstream vision models that are sensitive to illumination.
\begin{table}
    \centering
    \caption{Lighting direction preservation evaluated using scale-invariant Mean Squared Error (Si-MSE) and scale-invariant L$_2$ distance (Si-L$_2$) between the original and anonymized faces. Lower values ($\downarrow$) indicate better preservation. Best and second-best results are highlighted in blue and red respectively.}
    \label{tab:lighting-comparison}
    \begin{tabular}{lcc}
    \toprule
        Method & Si-MSE ($\downarrow$) & Si-L$_2$ ($\downarrow$) \\ \midrule
        AIDPro \cite{wang2025beyond} & 0.0314 $\pm$ 0.0122 & 0.0131 $\pm$ 0.0073 \\ 
        AnonyGAN \cite{dall2022graph}  & 0.0359 $\pm$ 0.0205 & 0.0146 $\pm$ 0.0081 \\
        CIAGAN \cite{maximov2020ciagan} & 0.0792 $\pm$ 0.0392 & 0.0185 $\pm$ 0.0095 \\
        DeepPrivacy \cite{hukkelaas2019deepprivacy} & 0.0549 $\pm$ 0.0355 & 0.0163 $\pm$ 0.0095 \\
        DeepPrivacy2 \cite{hukkelaas2022deepprivacy2} & 0.0838 $\pm$ 0.0516 & 0.0211 $\pm$ 0.0124 \\
        FALCO \cite{barattin2023attribute} & 0.0654 $\pm$ 0.0386 & 0.0163 $\pm$ 0.0088 \\
        $\mathrm{G^2}$Face~\cite{yang2024g} & \secondbest{0.0179 $\pm$ 0.0070} & \secondbest{0.0110 $\pm$ 0.0067} \\
        RiDDLE \cite{li2023riddle} & 0.0531 $\pm$ 0.0247 & 0.0185 $\pm$ 0.0083 \\
        \midrule
        Our & \best{0.0137 $\pm$ 0.0237} & \best{0.0100 $\pm$ 0.0085} \\
        \bottomrule 
    \end{tabular}
\end{table}

Finally, Table~\ref{tab:skin-color-results} compares the methods in terms of skin color preservation using pixelwise $L^*a^*b^*$ differences and $\Delta E_{2000}$ perceptual distance. Our method achieves one of the lowest mean $\Delta E$ values, nearly matching G$^2$Face, which obtains the best overall color consistency. Importantly, our method shows low variance across samples, indicating stable and consistent color reproduction. This is essential to prevent demographic bias amplification and to maintain downstream fairness in tasks involving skin-tone-sensitive analysis.
\begin{table}
    \centering
    \caption{Skin color preservation evaluated using SiMSE, lightness error (MSE-L), chroma error (MSE-ab), and perceptual difference ($\Delta E_{2000}$) between original and anonymized faces. Lower values indicate better color consistency. Best and second-best results are highlighted in blue and red respectively.}
    \label{tab:skin-color-results}
    \resizebox{\columnwidth}{!}{\begin{tabular}{lcccc}
    \toprule
        Method & Si-MSE & MSE-L & MSE-ab & $\Delta E$ \\ \midrule
        AIDPro \cite{wang2025beyond} & 3.60$\times10^{-7}$ & 0.1241 $\pm$ 0.1316 & 0.1426 $\pm$ 0.1128 & 10.30 $\pm$ 9.46 \\ 
        AnonyGAN \cite{dall2022graph} & 3.19$\times 10^{-7}$ & \secondbest{0.1053 $\pm$ 0.1083} & 0.1269 $\pm$ 0.0945 & 9.03 $\pm$ 7.84 \\
        CIAGAN \cite{maximov2020ciagan} & 6.82$\times 10^{-7}$ & 0.2153 $\pm$ 0.2320 & 0.2289 $\pm$ 0.1976 & 17.26 $\pm$ 16.61 \\
        DeepPrivacy \cite{hukkelaas2019deepprivacy} & 4.76$\times 10^{-7}$ & 0.1482 $\pm$ 0.1675 & 0.1710 $\pm$ 0.1446 & 12.27 $\pm$ 12.06 \\
        DeepPrivacy2 \cite{hukkelaas2022deepprivacy2} & 6.40$\times 10^{-7}$ & 0.2087 $\pm$ 0.2353 & 0.2226 $\pm$ 0.2010 & 16.73 $\pm$ 16.87 \\
        FALCO \cite{barattin2023attribute} & 4.89$\times 10^{-7}$ & 0.1605 $\pm$ 0.1457 & 0.1772 $\pm$ 0.1264 & 13.08 $\pm$ 10.50 \\
        G$^2$Face \cite{yang2024g} & \best{2.63$\times10^{-7}$} & \best{0.0946 $\pm$ 0.1310} & \best{0.1077 $\pm$ 0.1134} & \best{7.83 $\pm$ 9.46} \\ 
        RiDDLE \cite{li2023riddle} & 5.07$\times 10^{-7}$ & 0.1737 $\pm$ 0.1504 & 0.1941 $\pm$ 0.1306 & 14.13 $\pm$ 10.83 \\ \midrule
        Our & \secondbest{3.14 $\times10^{-7}$} & 0.1117 $\pm$ 0.1555 & \secondbest{0.1163 $\pm$ 0.1362} & \secondbest{8.93 $\pm$ 11.32} \\ 
        \bottomrule
    \end{tabular}}
\end{table}

\begin{figure}
    \centering
    \includegraphics[width=\columnwidth]{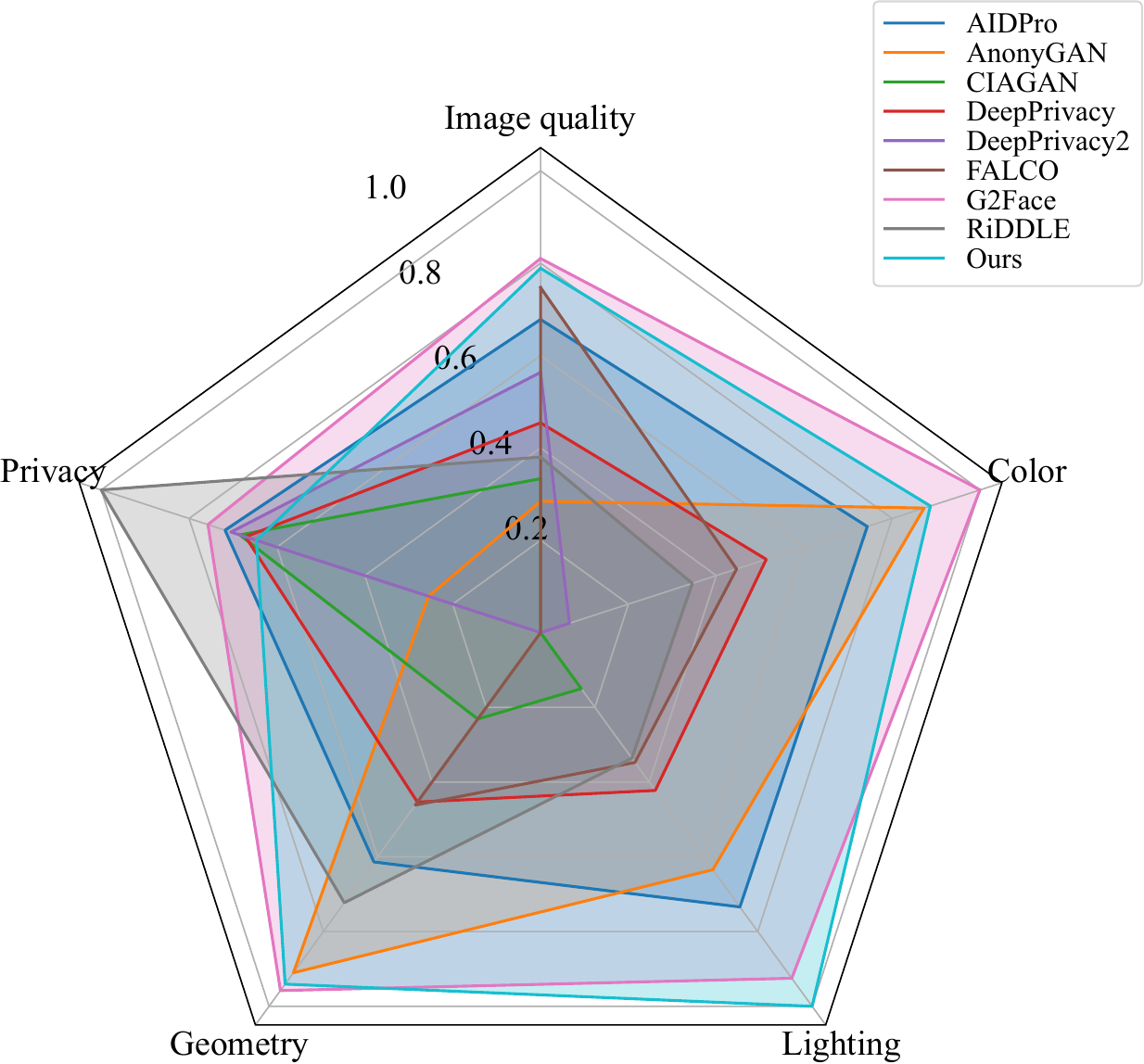}
    \caption{Aggregated performance across five evaluation dimensions: image quality, privacy, geometry, lighting, and color. Each axis represents the normalized average score for a dimension, where higher values indicate better performance.}
    \label{fig:radar-plot}
\end{figure}

Overall, the results across all tables demonstrate that our method strikes a favorable balance between image quality, privacy protection, and utility preservation. It achieves near state-of-the-art results in every metric, with particularly strong performance in expression coherence and lighting preservation — two dimensions often neglected by existing anonymization approaches. These findings confirm the effectiveness of our design choices and their suitability for real-world applications where visual realism, privacy, and downstream task usability must be simultaneously guaranteed. 
A global view of this trade-off is provided in Fig.~\ref{fig:radar-plot}, where the aggregated scores across the five evaluation dimensions are summarized in a radar plot, highlighting the overall superiority and balanced performance of our approach compared to competing methods.

\subsection{Ablation study}
In this section, we analyze the contribution of the main components of our anonymization framework, namely the facial landmark alignment, the relighting strategy, and the Laplacian pyramid refinement. Table~\ref{tab:ablation-result} reports the quantitative results obtained by progressively enabling these modules and measuring their effect on image fidelity, privacy protection, and photometric consistency.

The baseline configuration, which excludes landmark alignment, relighting, and Laplacian refinement, corresponds to a simplified version similar to DeepPrivacy~\cite{hukkelaas2019deepprivacy}. This baseline achieves moderate image quality (FID$_{\mathrm{ImageNet}}$=32.50, FID$_{\mathrm{CLIP}}$=21.10) and satisfactory privacy protection, but the resulting faces exhibit limited expression coherence and poor lighting consistency (SiMSE=0.0549). Introducing facial landmark alignment notably improves perceptual quality and facial geometry, reducing the FID scores by over 70\% and increasing expression accuracy from 53.29\% to 72.40\%. The detection rate also rises to 99.35\%, confirming that alignment ensures better face structure and stability during generation. Replacing standard relighting with the SfSNet-based relighting module further enhances expression accuracy (74.62\%) and pose estimation, but the FID scores slightly worsen due to the coarse reflectance estimation provided by SfSNet. This step highlights the importance of accurate illumination modeling for balancing realism and privacy. Switching to Intrinsic Decomposition-based relighting results in the best image fidelity among all variants (FID$_{\mathrm{ImageNet}}$=7.50, FID$_{\mathrm{CLIP}}$=6.02), along with superior lighting and color consistency (SiMSE=0.0208, $\Delta E$=9.36). The improvement confirms that a simple Retinex-based intrinsic decomposition can effectively approximate shading transfer at a fraction of the computational cost of deep relighting networks. This result highlights an additional advantage of our design: the relighting step adds negligible overhead, making the proposed pipeline suitable for real-time or large-scale anonymization scenarios without sacrificing photometric accuracy.

The method effectively disentangles shading and albedo components, preserving both facial geometry and global illumination patterns. Finally, the inclusion of the Laplacian pyramid refinement leads to the most balanced configuration overall, achieving strong visual realism, improved expression accuracy (75.00\%), and the lowest lighting and color errors (SiMSE=0.0137, $\Delta E$=8.93). This refinement step sharpens fine-scale details, mitigates over-smoothing, and harmonizes relight regions with the background.  

\begin{table*}
    \centering
    \caption{Ablation study. Effect of facial landmark alignment, relighting method, and Laplacian pyramid refinement on image fidelity, recognition, and lighting and color consistency metrics. The best resultas are highlited in blue and red respectively.}
    \label{tab:ablation-result}
    \resizebox{\textwidth}{!}{\begin{tabular}{lll|ccccccccc}
    \toprule
         \multirow{2}{*}{68 Landmarks} & \multirow{2}{*}{Relighting} & \multirow{2}{*}{Lapl. Pyr.} & \multirow{2}{*}{FID$_{\mathrm{ImageNet}}$} & \multirow{2}{*}{FID$_{\mathrm{CLIP}}$} & \multirow{2}{*}{Face det.} & \multicolumn{2}{c}{Face re-identification} & \multirow{2}{*}{Pose estim.} & \multirow{2}{*}{Facial expr.} & \multirow{2}{*}{Light. (SiMSE)} & \multirow{2}{*}{Skin color ($\Delta E$)} \\
          & & & & & & VGGFace2 & CASIA-WebFace \\ \midrule
          \xmark & \xmark & \xmark & 32.50 & 21.10 & 97.24 & \best{0.1917} & \best{0.2170} & 0.0520 & 53.29 & 0.0549 & 12.27 \\
         \checkmark & \xmark & \xmark & \secondbest{9.85} & \secondbest{7.20} & \best{99.35} & 0.2193 & 0.2237 & 0.0360 & 72.40 & 0.0430 & 9.74 \\
         \checkmark & SfSNet & \checkmark & 30.21 & 17.52 & 95.33 & 0.4641 & 0.4671 & \secondbest{0.0339} & 74.62 & 0.0363 & 10.71 \\
         \checkmark & Intr. Decom. & \xmark & \best{7.50} & \best{6.02} & \secondbest{98.94} & 0.2331 & 0.2366 & 0.0392 & \secondbest{74.77} & \secondbest{0.0208} & \secondbest{9.36} \\
         \checkmark & Intr. Decom. & \checkmark & 10.54 & 8.27 & 98.04 & \secondbest{0.2125} & \secondbest{0.2184} & \best{0.0334} & \best{75.00} & \best{0.0137} & \best{8.93} \\ \bottomrule
    \end{tabular}}
\end{table*}

%========================================================
\section{Conclusions}
\label{sec:conclusions}
In this work, we introduced a feature-preserving face anonymization framework that explicitly maintains three critical aspects often overlooked in current anonymization systems: facial expression, lighting consistency, and skin color fidelity. Building upon DeepPrivacy, our approach integrates dense facial landmarks to better retain expression details and introduces lightweight photometric post-processing modules based on intrinsic image decomposition and color transfer in the YCbCr space. Comprehensive experiments on the CelebA-HQ dataset demonstrate that our method achieves a favorable balance between privacy protection, image realism, and downstream usability, outperforming or matching state-of-the-art methods across multiple evaluation dimensions. In particular, it exhibits superior preservation of expression and lighting, two factors essential for affective computing, relighting, and fairness-aware computer vision. Beyond quantitative gains, our findings show that effective anonymization must conceal identity while preserving photometric and affective integrity. This paradigm shift is key for enabling anonymized datasets that remain valid for research and deployment in sensitive domains such as healthcare, driver monitoring, and social robotics. Future work will explore diffusion-based extensions and domain adaptation strategies to further generalize the proposed framework across diverse imaging conditions.

\section*{Acknowledgment}

The authors would like to thank Sofia Zonari for her helpful support to this project during her Master's degree stage at the Imaging and Vision lab. Financial support from ICSC – Centro Nazionale di Ricerca in High Performance Computing, Big Data and Quantum Computing, funded by European Union – NextGenerationEU.

\ifCLASSOPTIONcaptionsoff
  \newpage
\fi



\bibliographystyle{IEEEtran}



\begin{IEEEbiography}[{\includegraphics[width=1in,height=1.25in,clip,keepaspectratio]{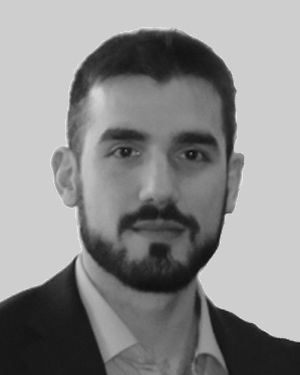}}]{Luigi Celona}
is currently an assistant professor at DISCo (Department of Informatics, Systems and Communication) of the University of Milano-Bicocca, Italy. In 2018 and 2014, he obtained respectively the PhD and the MSc degree in Computer Science at DISCo. In 2011, he obtained the BSc degree in Computer Science from the University of Messina. His current research interests focus on image analysis and classification, machine learning and face analysis.
\end{IEEEbiography}

\begin{IEEEbiography}[{\includegraphics[width=1in,height=1.25in,clip,keepaspectratio]{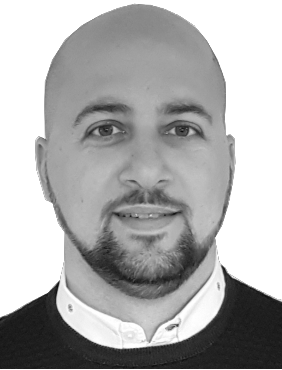}}]{Simone Bianco} is currently an Associate Professor of computer science with the University of Milano-Bicocca, Milan, Italy, holder of the Italian National Academic Qualification as a Full Professor of computer engineering (09/H1) and computer science (01/B1). He is on Stanford University's World Ranking Scientists List for his achievements in Artificial Intelligence and Image Processing. His teaching and research interests include computer vision, artificial intelligence, machine learning, optimization algorithms applied in multimodal, and multimedia applications. He is also a R\&D Manager of the University of Milano-Bicocca spin-off Imaging and Vision Solutions, and Member of European Laboratory for Learning and Intelligent Systems.
\end{IEEEbiography}

\begin{IEEEbiography}[{\includegraphics[width=1in,height=1.25in,clip,keepaspectratio]{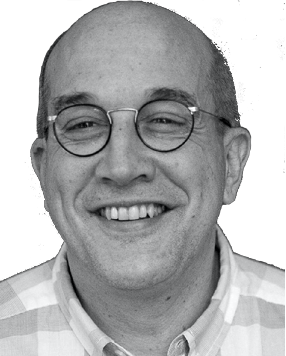}}]{Raimondo Schettini} is currently a Full Professor
with the University of Milano-Bicocca, Milan, Italy, where he leads the Imaging and Vision Lab. Since 1987, he has been associated with the Italian National Research Council, where he led the Color Imaging Lab from 1990 to 2002. He is a team leader in several research projects, some of them supported by companies. He authored or coauthored more than 400 refereed papers and 12 patents about color imaging, image processing, analysis, and classification, image, and video understanding and retrieval. He supervised more than ten Ph.D. students. He is the Chair of several international conferences and workshops and he is the Editor in Chief of the MDPI Journal of Imaging. He is on Stanford University's World Ranking Scientists List for his achievements in artificial intelligence and image processing. He is a Fellow of the International Association of Pattern Recognition for his contributions to pattern recognition research and color image analysis, and Fellow of Asia-Pacific Artificial Intelligence Association. Raimondo Schettini is also the Chief Technical Officer of the University of Milano-Bicocca spin-off Imaging and Vision Solutions, Member of European Laboratory for Learning and Intelligent Systems, and Member of the advisory board of the international AIQT Foundation, an international competence platform for the active public and private exchange of experience in the fields of artificial intelligence and quantum technology.
\end{IEEEbiography}

%\vfill


\end{document}